\pdfoutput=1
\documentclass{article}
\usepackage{style/unites}
\usepackage{XCharter}
\usepackage[scaled=1.1]{zlmtt}

\usepackage[utf8]{inputenc}
\usepackage[T1]{fontenc}
\usepackage{microtype}

\usepackage{amsmath}
\usepackage{amssymb}
\usepackage{amsfonts}
\usepackage{amsthm}
\usepackage{mathtools}
\usepackage{mathrsfs}
\usepackage{physics}
\usepackage{braket}
\usepackage{slashed}
\usepackage{nicefrac}
\usepackage{textcomp}
\usepackage{dsfont}
\usepackage{bbm}
\usepackage{bm}

\usepackage{graphicx}
\usepackage{subcaption}
\usepackage[export]{adjustbox}
\usepackage{float}
\usepackage{booktabs}
\usepackage{dcolumn}
\newcolumntype{d}[1]{D{.}{.}{#1}}
\usepackage{bigstrut, tabularx, multirow, makecell, diagbox}
\usepackage{colortbl}
\usepackage{tabularray}
\UseTblrLibrary{booktabs}
\usepackage{threeparttable}
\usepackage{tablefootnote}
\usepackage{fontawesome5}

\usepackage{placeins}
\usepackage{caption}
\usepackage{footnote}
\usepackage{enumitem}
\usepackage{multicol}
\usepackage{xspace}
\usepackage{titletoc}
\usepackage{titlesec}
\usepackage[bottom]{footmisc}
\usepackage{setspace}

\usepackage{wrapfig}
\usepackage{tikz}
\usepackage{quantikz}
\usepackage{dashbox}
\usepackage{mdframed}
\usepackage{marvosym}
\usepackage{pifont}
\usepackage{CJK}
\usepackage{url}

\usepackage[table,x11names]{xcolor}
\usepackage[most]{tcolorbox}
\tcbuselibrary{breakable}
\usetikzlibrary{decorations.pathreplacing, fit}

\definecolor{primaryblue}{HTML}{0066CC}
\definecolor{accentcyan}{HTML}{00D4AA}
\definecolor{warmorange}{HTML}{FF6B35}
\definecolor{deepgray}{HTML}{2C3E50}
\definecolor{lightgray}{HTML}{F8F9FA}
\definecolor{gradientstart}{HTML}{667eea}
\definecolor{gradientend}{HTML}{764ba2}

\definecolor{citecolor}{HTML}{0071bc}
\definecolor{citeblue}{RGB}{0, 113, 188}
\definecolor{linkcolor}{HTML}{9A4D92}
\definecolor{firebrick}{rgb}{0.698,0.133,0.133}

\definecolor{paleviolet}{HTML}{E1EEFC}
\definecolor{CarolinaUltraLight}{HTML}{E7F4FC}
\definecolor{lightgrey}{RGB}{247, 247, 247}
\definecolor{shadecolor}{HTML}{EFEFEF}
\definecolor{lightyellow}{rgb}{1.0, 0.95, 0.7}
\definecolor{lightblue}{rgb}{0.90, 0.95, 1.0}
\definecolor{light-gray}{gray}{0.95}

\definecolor{darkgrey}{rgb}{0.5, 0.5, 0.5}
\definecolor{darkgreen}{rgb}{0, 0.5, 0}
\definecolor{mydarkblue}{rgb}{0,0.08,0.45}
\definecolor{mydarkblue2}{rgb}{0.133, 0.133, 0.698}
\definecolor{echodrk}{HTML}{0099cc}
\definecolor{mymauve}{rgb}{0.58,0,0.82}
\definecolor{midnightblue}{rgb}{0.1,0.1,0.44}
\definecolor{oxfordblue}{rgb}{0.0,0.13,0.28}
\definecolor{prussianblue}{rgb}{0.0,0.19,0.33}
\definecolor{coolteal}{rgb}{0, 0.45, 0.45}
\definecolor{olive}{rgb}{0.1, 0.3, 0}
\definecolor{mypurple}{rgb}{0.5,0,0.5}
\definecolor{almond}{rgb}{0.94, 0.87, 0.8}

\definecolor{blue_ampEncoding}{HTML}{DAE8FC}
\definecolor{green_encoder}{HTML}{D5E8D4}
\definecolor{purple_decoder}{HTML}{E1D5E7}
\definecolor{yellow_measure}{HTML}{FFF2CC}
\definecolor{gray_block}{HTML}{F5F5F5}
\definecolor{pink_dru}{HTML}{FAD9D5}
\definecolor{orange_v}{HTML}{FAD7AC}

\definecolor{colorA}{rgb}{1,0,0}
\definecolor{colorB}{rgb}{0,0.3,1}
\definecolor{colorC}{rgb}{0.9,0.8,0.2}
\definecolor{colorD}{rgb}{0,0.65,0}
\definecolor{lesslightgray}{rgb}{0.5,0.5,0.5}
\definecolor{fundamental}{RGB}{55, 110, 111}
\definecolor{Gred}{RGB}{219, 50, 54}
\definecolor{ToCgreen}{RGB}{0, 128, 0}
\definecolor{Sepia}{RGB}{112, 66, 20}
\definecolor{Dblue}{rgb}{0,0.08,0.45}
\definecolor{Blue}{rgb}{0, 0, 0.8}
\definecolor{blue}{rgb}{0,0,1}
\definecolor{UNCblue!10}{rgb}{0.84,0.91,0.98}
\definecolor{RowAlt}{rgb}{0.98,0.98,0.99}

\definecolor{CarolinaBlue}{HTML}{7BAFD4}        %
\definecolor{CarolinaLightBlue}{HTML}{B3D4E5}   %
\definecolor{CarolinaUltraLight}{HTML}{E8F4F8}  %
\definecolor{CarolinaText}{HTML}{1C2B33}        %

\usepackage[pagebackref=true,breaklinks=true,colorlinks,hyperfootnotes=false]{hyperref}
\hypersetup{
  colorlinks,
  citecolor=citeblue,
  linkcolor=firebrick,
  urlcolor=firebrick
}
\usepackage[nameinlink,capitalize,noabbrev]{cleveref}

\titlespacing\section{0pt}{4pt plus 4pt minus 2pt}{-2pt plus 2pt minus 2pt}
\titlespacing\subsection{0pt}{2pt plus 4pt minus 2pt}{-2pt plus 2pt minus 2pt}
\titlespacing\subsubsection{0pt}{2pt plus 4pt minus 2pt}{-2pt plus 2pt minus 2pt}

\makeatletter
\def\th@remark{%
  \thm@headfont{\bfseries}%
  \normalfont %
  \thm@preskip\topsep \divide\thm@preskip\tw@
  \thm@postskip\thm@preskip
}
\makeatother

\theoremstyle{definition}

\tcolorboxenvironment{theorem}{
  breakable,
  colback=black!10,
  colframe=white,
  width=\linewidth, 
  enlarge left by=0pt,
  enlarge right by=0pt,
  boxsep=5pt,
  boxrule=0pt,
  left=0pt,right=0pt,top=0pt,bottom=0pt,
  arc=8pt,
  before skip=\topsep,
  after skip=\topsep
}

\tcolorboxenvironment{lemma}{
  breakable,
  colback=black!10,
  colframe=white,
  width=\linewidth,
  enlarge left by=0pt,
  enlarge right by=0pt,
  boxsep=5pt,
  boxrule=0pt,
  left=0pt,right=0pt,top=0pt,bottom=0pt,
  arc=8pt,
  before skip=\topsep,
  after skip=\topsep
}

\tcolorboxenvironment{corollary}{
  breakable,
  colback=black!10,
  colframe=white,
  width=\linewidth,
  enlarge left by=0pt,
  enlarge right by=0pt,
  boxsep=5pt,
  boxrule=0pt,
  left=0pt,right=0pt,top=0pt,bottom=0pt,
  arc=8pt,
  before skip=\topsep,
  after skip=\topsep
}

\tcolorboxenvironment{proposition}{
  breakable,
  colback=black!10,
  colframe=white,
  width=\linewidth,
  enlarge left by=0pt,
  enlarge right by=0pt,
  boxsep=5pt,
  boxrule=0pt,
  left=0pt,right=0pt,top=0pt,bottom=0pt,
  arc=8pt,
  before skip=\topsep,
  after skip=\topsep
}

\tcolorboxenvironment{definition}{
  breakable,
  colback=black!10,
  colframe=white,
  width=\linewidth,
  enlarge left by=0pt,
  enlarge right by=0pt,
  boxsep=5pt,
  boxrule=0pt,
  left=0pt,right=0pt,top=0pt,bottom=0pt,
  arc=8pt,
  before skip=\topsep,
  after skip=\topsep
}

\tcolorboxenvironment{assumption}{
  breakable,
  colback=black!10,
  colframe=white,
  width=\linewidth,
  enlarge left by=0pt,
  enlarge right by=0pt,
  boxsep=5pt,
  boxrule=0pt,
  left=0pt,right=0pt,top=0pt,bottom=0pt,
  arc=8pt,
  before skip=\topsep,
  after skip=\topsep
}

\tcolorboxenvironment{claim}{
  breakable,
  colback=black!10,
  colframe=white,
  width=\linewidth,
  enlarge left by=0pt,
  enlarge right by=0pt,
  boxsep=5pt,
  boxrule=0pt,
  left=0pt,right=0pt,top=0pt,bottom=0pt,
  arc=8pt,
  before skip=\topsep,
  after skip=\topsep
}

\tcolorboxenvironment{problem}{
  breakable,
  colback=black!10,
  colframe=white,
  width=\linewidth,
  enlarge left by=0pt,
  enlarge right by=0pt,
  boxsep=5pt,
  boxrule=0pt,
  left=0pt,right=0pt,top=0pt,bottom=0pt,
  arc=8pt,
  before skip=\topsep,
  after skip=\topsep
}

\tcolorboxenvironment{question}{
  breakable,
  colback=black!10,
  colframe=white,
  width=\linewidth,
  enlarge left by=0pt,
  enlarge right by=0pt,
  boxsep=5pt,
  boxrule=0pt,
  left=0pt,right=0pt,top=0pt,bottom=0pt,
  arc=8pt,
  before skip=\topsep,
  after skip=\topsep
}

\newtcolorbox{titleblock}{
  enhanced,
  frame hidden,
  colback=CarolinaUltraLight,
  colframe=CarolinaUltraLight,
  boxrule=0pt,
  arc=10pt,
  left=14pt,
  right=14pt,
  top=14pt,
  bottom=14pt,
  width=\linewidth,
  before skip=12pt plus 4pt,
  after skip=12pt plus 4pt,
  grow to left by=1.5pt,
  grow to right by=1.5pt,
  before upper={
    \setlength{\parindent}{0cm}
    \setlength{\parskip}{0.5cm}
  }
}

\crefname{theorem}{Theorem}{Theorems}
\crefname{proposition}{Proposition}{Propositions}
\crefname{lemma}{Lemma}{Lemmas}
\crefname{corollary}{Corollary}{Corollaries}
\crefname{definition}{Definition}{Definitions}
\crefname{assumption}{Assumption}{Assumptions}
\crefname{remark}{Remark}{Remarks}
\crefname{problem}{Problem}{Problems}
\crefname{property}{Property}{property}
\crefname{question}{Question}{Questions}

\numberwithin{equation}{section}
\numberwithin{theorem}{section}
\numberwithin{proposition}{section}
\numberwithin{definition}{section}
\numberwithin{lemma}{section}
\numberwithin{assumption}{section}
\numberwithin{remark}{section}

\newcommand{\best}[1]{\mathbf{#1}}

\def\1{\bm{1}}

\makeatletter
\let\save@mathaccent\mathaccent
\newcommand*\if@single[3]{%
    \setbox0\hbox{${\mathaccent"0362{#1}}^H$}%
    \setbox2\hbox{${\mathaccent"0362{\kern0pt#1}}^H$}%
    \ifdim\ht0=\ht2 #3\else #2\fi
}
\newcommand*\rel@kern[1]{\kern#1\dimexpr\macc@kerna}
\newcommand*\widebar[1]{\@ifnextchar^{{\wide@bar{#1}{0}}}{\wide@bar{#1}{1}}}
\newcommand*\wide@bar[2]{\if@single{#1}{\wide@bar@{#1}{#2}{1}}{\wide@bar@{#1}{#2}{2}}}
\newcommand*\wide@bar@[3]{%
    \begingroup
    \def\mathaccent##1##2{%
        \let\mathaccent\save@mathaccent
        \if#32 \let\macc@nucleus\first@char \fi
        \setbox\z@\hbox{$\macc@style{\macc@nucleus}_{}$}%
        \setbox\tw@\hbox{$\macc@style{\macc@nucleus}{}_{}$}%
        \dimen@\wd\tw@
        \advance\dimen@-\wd\z@
        \divide\dimen@ 3
        \@tempdima\wd\tw@
        \advance\@tempdima-\scriptspace
        \divide\@tempdima 10
        \advance\dimen@-\@tempdima
        \ifdim\dimen@>\z@ \dimen@0pt\fi
        \rel@kern{0.6}\kern-\dimen@
        \if#31
        \overline{\rel@kern{-0.6}\kern\dimen@\macc@nucleus\rel@kern{0.4}\kern\dimen@}%
        \advance\dimen@0.4\dimexpr\macc@kerna
        \let\final@kern#2%
        \ifdim\dimen@<\z@ \let\final@kern1\fi
        \if\final@kern1 \kern-\dimen@\fi
        \else
        \overline{\rel@kern{-0.6}\kern\dimen@#1}%
        \fi
    }%
    \macc@depth\@ne
    \let\math@bgroup\@empty \let\math@egroup\macc@set@skewchar
    \mathsurround\z@ \frozen@everymath{\mathgroup\macc@group\relax}%
    \macc@set@skewchar\relax
    \let\mathaccentV\macc@nested@a
    \if#31
    \macc@nested@a\relax111{#1}%
    \else
    \def\gobble@till@marker##1\endmarker{}%
    \futurelet\first@char\gobble@till@marker#1\endmarker
    \ifcat\noexpand\first@char A\else
    \def\first@char{}%
    \fi
    \macc@nested@a\relax111{\first@char}%
    \fi
    \endgroup
    }
\makeatother

\DeclareMathAlphabet{\mathsfit}{\encodingdefault}{\sfdefault}{m}{sl}
\SetMathAlphabet{\mathsfit}{bold}{\encodingdefault}{\sfdefault}{bx}{n}

\renewcommand{\arraystretch}{1.15}
\usepackage{algpseudocode}

\makeatletter
\newcommand{\clearwrap}{\WFclear}
\makeatother
\definecolor{darkgreen}{RGB}{0,80,60}
\definecolor{darkrose}{RGB}{90,0,120}
\definecolor{darkblue}{RGB}{0,60,100}
\definecolor{darkorange}{RGB}{170,80,20}
\definecolor{fluyellow}{RGB}{255,245,80}
\newcolumntype{H}{>{\columncolor{fluyellow!25}}r}
\renewcommand{\best}[1]{\cellcolor{fluyellow!25}\textbf{#1}}

\newcommand{\cmark}{\ding{51}}
\newcommand{\xmark}{\ding{55}}
\newcommand{\pmark}{$\triangle$}

\hypersetup{
  pdftitle={AgentOrch: How Close Can Global Orchestration Come to Step-by-Step Interaction?},
  pdfauthor={Zijie Liu, Qi Yang Huang, Jinhao Duan, Yueh Che Lee, Thomas A. Joseph, Zayne Turner, Adam Seligman, Tianlong Chen, Zhaozhuo Xu}
}

\begin{document}

\makeatletter
\def\blfootnote{\gdef\@thefnmark{}\@footnotetext}
\makeatother

\makeatletter
\pagestyle{fancy}
\fancyhf{}
\renewcommand{\headrulewidth}{1pt}
\chead{\fontsize{8}{9}\selectfont\bfseries\textsc{AgentOrch}: Global Orchestration and Step-by-Step Interaction}
\cfoot{\thepage}
\thispagestyle{fancy}
\makeatother

\makeatletter
\def\icmldate#1{\gdef\@icmldate{#1}}
\icmldate{October 23, 2026}
\makeatother

\makeatletter
\fancypagestyle{fancytitlepage}{
  \fancyhead{}
  \lhead{\smash{%
    \includegraphics[height=0.8cm]{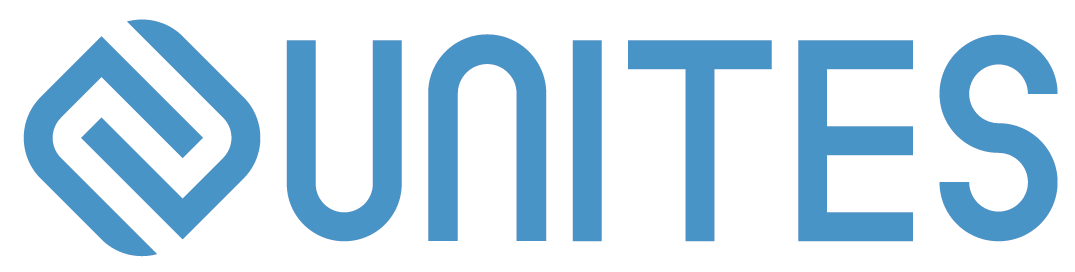}%
    \hspace{0.5cm}%
    \raisebox{0.15cm}{\includegraphics[height=0.5cm]{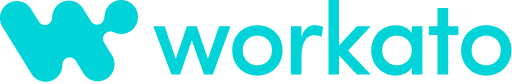}}%
  }}
  \rhead{\fontsize{8}{9}\selectfont\itshape \@icmldate}
  \cfoot{}
}
\makeatother

\thispagestyle{fancytitlepage}

\vspace*{0.5em}

\noindent
\begin{titleblock}
    {\setlength{\parskip}{0cm}
     \raggedright
     {\setstretch{1.2}
      \LARGE\bfseries
\textsc{AgentOrch}: How Close Can Global Orchestration Come to Step-by-Step Interaction?

      \par}
    }
    \vskip 0.2cm
    
\begin{icmlauthorlist}
\mbox{Zijie Liu$^{1,\textrm{\Letter}}$},
\mbox{Qi Yang Huang$^{2}$},
\mbox{Jinhao Duan$^{1}$},
\mbox{Yueh Che Lee$^{2}$},
\mbox{Thomas A. Joseph$^{2}$},
\mbox{Zayne Turner$^{2}$},
\mbox{Adam Seligman$^{2}$},
\mbox{Tianlong Chen$^{1}$},
and \mbox{Zhaozhuo Xu$^{2}$}
\end{icmlauthorlist}

\noindent $^{1}$University of North Carolina at Chapel Hill
\quad $^{2}$Workato\par

    \vskip 0.2cm
    
Enterprise workflow agents typically rely on step-by-step interaction (\textsc{Step-by-Step}): the agent executes an action, observes its result, and invokes a large language model (LLM) again to decide the next step. Global orchestration (\textsc{Global}) instead generates a complete executable workflow in advance and executes it without further LLM calls, substantially reducing inference overhead, but it remains less accurate than \textsc{Step-by-Step}. We study how close \textsc{Global} can come to \textsc{Step-by-Step} and what narrows the remaining gap. We introduce \textsc{AgentOrch}, a benchmark that directly compares global orchestration with step-by-step interaction on 650 enterprise SOPs and 3,220 runtime variants under matched tools, environments, and evaluation. \textsc{Step-by-Step} remains more accurate for every evaluated model, but the gap has shrunk by about 79\% from early models to the strongest recent model. We identify three routes to further narrowing this gap. Incremental workflow construction improves \textsc{Global} accuracy by an average of 9.98 percentage points (pp) without runtime execution feedback. Revising complete workflows with execution feedback improves all evaluated models, including 6.13 pp on held-out runtime variants. Targeted training on complete workflows improves accuracy by an average of 12.18 pp across all five trained models. Although a gap remains, global orchestration offers a promising path toward enterprise agents that deliver higher accuracy with fewer LLM calls.

\end{titleblock}

\blfootnote{%
$^{\textrm{\Letter}}$ Zijie Liu:
\href{mailto:jesseliu@cs.unc.edu}{\texttt{jesseliu@cs.unc.edu}}.
Codebase: \href{https://github.com/workato-devs/AgentOrch}{\texttt{github.com/workato-devs/AgentOrch}}.
}

\section{Introduction}

Enterprise workflows often span multiple software tools and services, with later operations depending on the outputs of earlier ones~\citep{li2026tool,wang2026mcp,lu2025toolsandbox,trivedi2024appworld,xie2024osworld,boisvert2024workarena++,drouin2024workarena}. Completing these workflows requires an LLM agent to select tools~\citep{qin2024toolllm,song2023restgpt,shen2023hugginggpt,schick2023toolformer} and coordinate their execution order, data dependencies, and control flow~\citep{liu2024agentbench,nath2025toolcomp,li2026tool}. We refer to this capability as \emph{workflow orchestration}.

\begin{figure}[htbp]
    \centering
    \includegraphics[width=\textwidth]{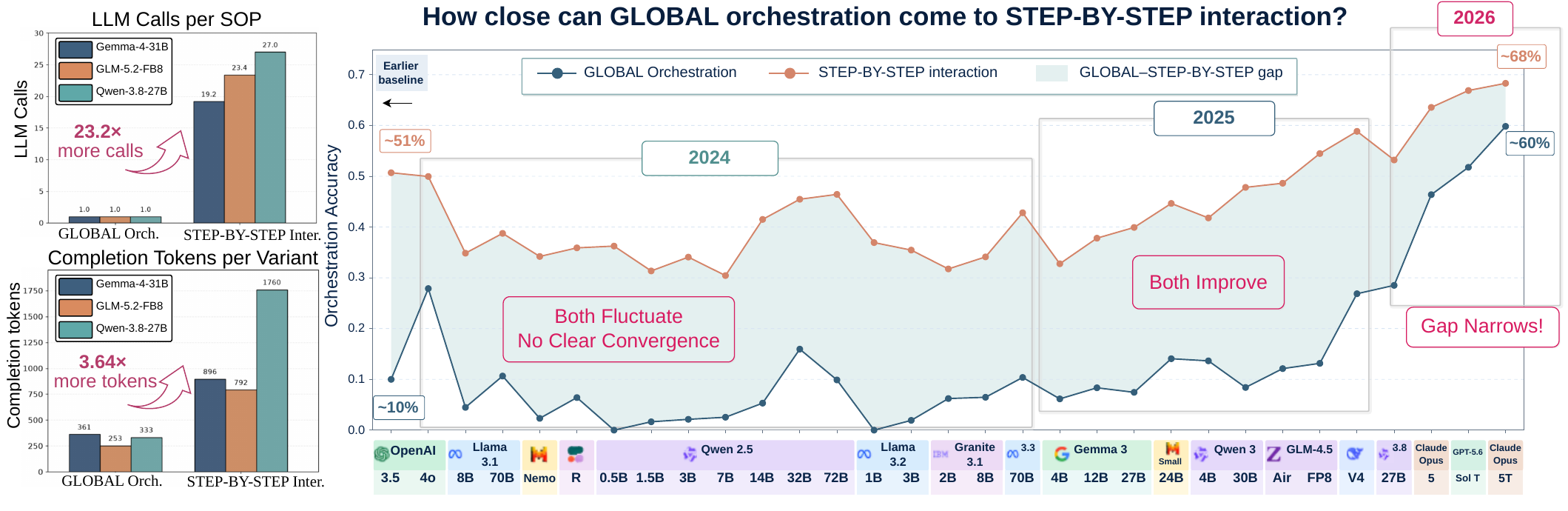}
    \vspace{-5mm}
\caption{\small 
\textbf{Global orchestration reduces inference overhead but remains less accurate than step-by-step interaction.}
\textbf{Left:} \textsc{Step-by-Step} uses on average 23.2$\times$ more LLM calls per SOP and 3.64$\times$ more completion tokens per runtime variant than \textsc{Global}.
\textbf{Right:} \textsc{Step-by-Step} achieves higher Orchestration Accuracy across all evaluated models, while the gap narrows to approximately 8 percentage points for Claude Opus 5T.
}
    \label{fig:teaser}
\end{figure}

Existing work commonly realizes workflow orchestration through step-by-step interaction (\textsc{Step-by-Step}): the model selects an action, observes its execution result, and then determines the next action~\citep{shi2024learning,yao2022react,zhou2024webarena,liu2024agentbench}. This interaction allows subsequent decisions to incorporate execution feedback~\citep{liu2024agentbench,shinn2023reflexion}, but requires repeated LLM inference throughout execution. Yet dependence on intermediate results does not inherently require the model itself to make a new decision after every action. Data dependencies, branching, loops, and other control flow can instead be specified before execution, allowing intermediate results to be consumed by a preconstructed workflow~\citep{liang2023code,singh2022progprompt,suris2023vipergpt}. We refer to this paradigm as \emph{global orchestration} (\textsc{Global}): the LLM constructs a complete executable workflow before observing concrete runtime inputs, after which the workflow executes without further LLM calls. As shown in Figure~\ref{fig:teaser}, across the three models in the cost panels, \textsc{Step-by-Step} makes 23.2 times as many LLM calls and generates 3.64 times as many completion tokens as \textsc{Global}, on average. As LLMs become more capable at planning and tool execution~\citep{kim2023llm,wang2023describe,zhou2023language}, it becomes increasingly important to understand how much orchestration can be completed before execution. Prior work has explored generating executable workflows or plans before execution~\citep{fan2025workflowllm,qiao2025benchmarking,zhang2025aflow}, but typically evaluates them in isolation rather than against step-by-step interaction on the same workflows, tools, execution environment, and criteria. We therefore ask:

\begin{mdframed}[
    userdefinedwidth=.99\linewidth,
    align=center,
    skipabove=3pt,
    skipbelow=3pt,
    innerleftmargin=6pt,
    innerbottommargin=6pt,
    innertopmargin=6pt,
    roundcorner=3pt,
    backgroundcolor=cyan!5,
    linecolor=gray
]
\noindent
\textbf{Question}: \textit{How close can Global Orchestration come to Step-by-Step Interaction?}
\end{mdframed}

We first make an initial comparison between \textsc{Global} and \textsc{Step-by-Step} across model generations, shown in Figure~\ref{fig:teaser}. \textsc{Step-by-Step} achieves higher Orchestration Accuracy for every evaluated model, but the magnitude of this advantage changes substantially across generations. From the earlier baselines through the 2024 and 2025 models, both settings improve without a consistent reduction in the gap. The 2026 models show clearer convergence, with gaps of 15.12 percentage points (pp) for GPT-5.6 Sol Thinking and 8.53 pp for Claude Opus 5 Thinking, compared with 40.67 pp for GPT-3.5 Turbo, a reduction of about 79\%. Thus, global orchestration has become substantially stronger and the gap narrows in the latest model generation, but \textsc{Step-by-Step} continues to retain an accuracy advantage across all evaluated models.

Then to systematically study this gap, we introduce \textsc{AgentOrch}, a benchmark grounded in 650 enterprise standard operating procedures (SOPs) and 3,220 runtime variants. Each SOP is associated with an executable reference workflow and multiple runtime variants, across which the same generated workflow is evaluated. Importantly, \textsc{AgentOrch} evaluates \textsc{Global} and \textsc{Step-by-Step} on the same SOPs, tool interfaces, execution environment, and evaluation criteria, enabling a direct and controlled comparison between the two orchestration protocols.

Using \textsc{AgentOrch}, we first examine what determines how close \textsc{Global} can come to \textsc{Step-by-Step}. Increasing model scale does not consistently narrow the gap, while newer model generations generally improve global orchestration and reasoning produces mixed effects. We then identify three routes to narrowing the gap, each drawing on progressively more signal. Incremental workflow construction improves \textsc{Global} Orchestration Accuracy by an average of 9.98 pp even without access to runtime execution feedback. We next revise the \textsc{Global} workflow using feedback from its own executions, without adding LLM calls at execution time. This improves Orchestration Accuracy for all eight evaluated models by an average of 6.18 pp, and by 6.13 pp on held-out runtime variants never used for feedback. Finally, we study whether stronger global orchestration can be internalized into model weights. We construct a separate SOP-grounded training set from 3,000 source recipes, with each sample pairing a task specification with a complete reference workflow. Targeted training improves Orchestration Accuracy by an average of 12.18 pp (6.84--19.88 pp) across all five trained models. Together, these results show that global orchestration can be substantially improved through incremental workflow construction, execution-feedback revision, and targeted training, while \textsc{Step-by-Step} continues to retain an accuracy advantage across the evaluated models.
Our contributions are summarized as follows:
\begin{itemize}[leftmargin=*]
    \item \textbf{Orchestration Benchmark.} We introduce \textsc{AgentOrch}, an execution-grounded benchmark comparing \textsc{Global} and \textsc{Step-by-Step} on 650 enterprise SOPs and 3,220 runtime variants under shared tools, environments, and evaluation criteria.

    \item \textbf{Test-time Orchestration Improvement.} We evaluate incremental workflow construction and execution-feedback revision as routes to stronger \textsc{Global} orchestration. Construction improves several models without runtime execution feedback; revision improves all eight evaluated models, including on held-out variants.

    \item \textbf{Training for Global Orchestration.} We train models on 3,000 separate source recipes to generate complete workflows. Taking the better of SFT and RL for each base model, Orchestration Accuracy improves by an average of 12.18 pp across all five trained models.
\end{itemize}

\begin{figure}[htbp]
    \centering
    \includegraphics[width=\textwidth]{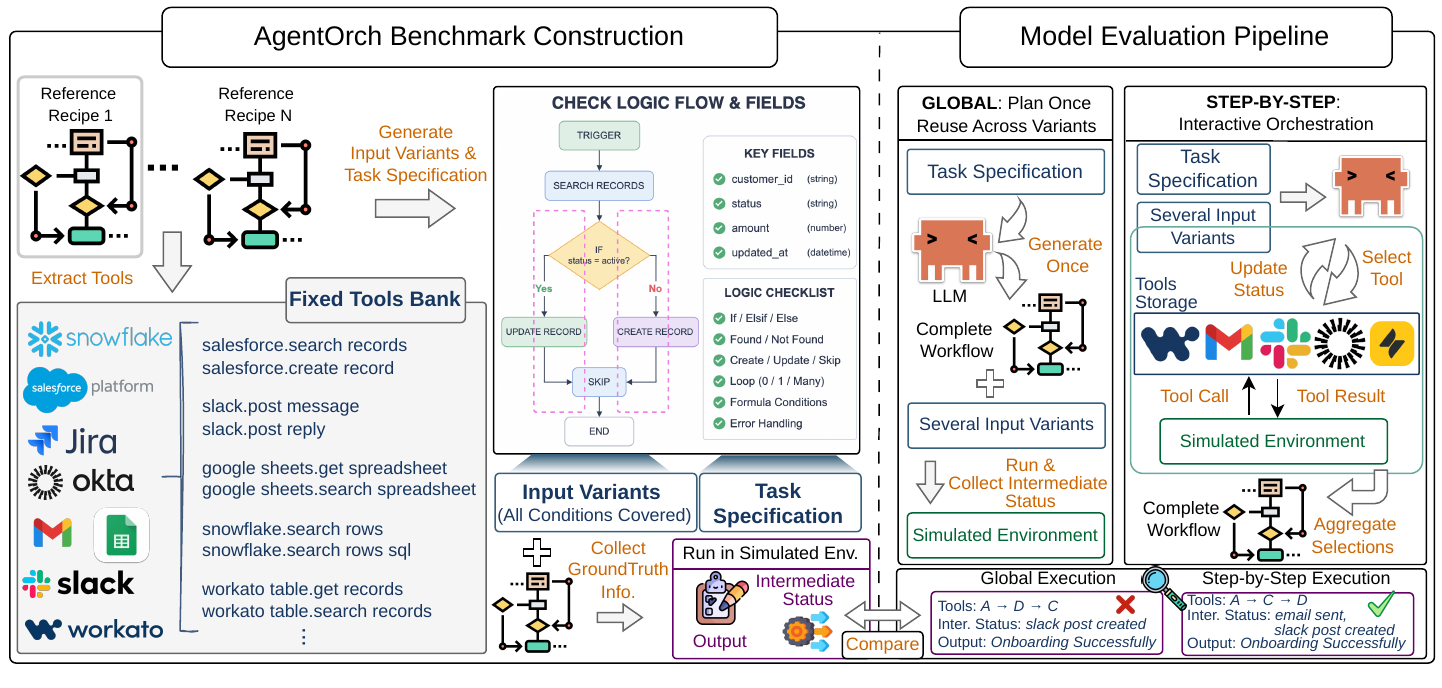}
    \caption{\small \textbf{Overview of the \textsc{AgentOrch} benchmark construction and model evaluation pipeline.}
    \textbf{Left:} Reference recipes define the benchmark SOPs. We extract their tool operations, map them to a fixed tool bank, construct task specifications and runtime variants, and validate the resulting workflow logic and fields. Reference executions provide the behavior used for evaluation.
    \textbf{Right:} Under \textsc{Global}, the model generates one complete workflow before execution and reuses it across runtime variants. Under \textsc{Step-by-Step}, the model selects tool calls sequentially using intermediate execution results. Both protocols are executed in the same simulated environment and evaluated against the corresponding reference behavior.}
    
    \label{fig:benchmark_pipeline}
\end{figure}

\section{Related Work}
\textbf{Interactive software-use benchmarks.}
A broad line of work evaluates language agents through step-by-step interaction with software environments. WebShop~\citep{yao2022webshop}, Mind2Web~\citep{deng2023mind2web}, WebArena~\citep{zhou2024webarena}, and VisualWebArena~\citep{koh2024visualwebarena} study realistic websites, while WorkArena~\citep{drouin2024workarena} and WorkArena++~\citep{boisvert2024workarena++} target enterprise knowledge work. OSWorld~\citep{xie2024osworld}, AndroidWorld~\citep{rawles2025androidworld}, AgentBench~\citep{liu2024agentbench}, SWE-bench~\citep{jimenez2024swe}, and TheAgentCompany~\citep{xu2026theagentcompany} extend evaluation to desktop, mobile, coding, and simulated workplace settings. These benchmarks generally allow agents to repeatedly act on and adapt to execution feedback. Our work instead asks how close global orchestration can come to step-by-step interaction by directly evaluating both paradigms on the same enterprise workflows.

\textbf{Tool-use and API-agent benchmarks.}
API-Bank~\citep{li2023api}, ToolLLM~\citep{qin2024toolllm}, Gorilla~\citep{patil2024gorilla}, ToolAlpaca~\citep{tang2023toolalpaca}, RestGPT~\citep{song2023restgpt}, StableToolBench~\citep{guo2024stabletoolbench}, TaskBench~\citep{shen2024taskbench}, UltraTool~\citep{huang2024planning}, $\tau$-bench~\citep{yao2024tau}, and WorkBench~\citep{styles2024workbench} evaluate tool retrieval, action selection, argument prediction, planning, multi-step execution, and user-agent interaction. While these capabilities are central to enterprise orchestration, they are typically evaluated under a fixed execution protocol. In contrast, \textsc{AgentOrch} evaluates \textsc{Global} and \textsc{Step-by-Step} under matched tasks, tools, execution environments, and evaluation criteria, enabling a direct comparison of the two orchestration paradigms.

\textbf{Summary.}
Table~\ref{tab:benchmark_comparison} summarizes these differences. Existing benchmarks typically evaluate either step-by-step interaction or complete workflow generation, whereas \textsc{AgentOrch} supports both under the same executable setting with reference workflows, state grading, and side-effect evaluation. This enables direct comparison between \textsc{Global} and \textsc{Step-by-Step}. Additional related work is discussed in Appendix~\ref{app:rel_wok}.

\begin{table*}[t]
\centering
\small
\setlength{\tabcolsep}{4.5pt}
\renewcommand{\arraystretch}{1.12}

\caption{\small 
Comparison of evaluation settings across agent and workflow benchmarks.
\textbf{E2E Exec.} indicates whether generated actions or workflows are executed end-to-end in an environment.
\textbf{\textsc{Step-by-Step}} indicates whether the model can make new orchestration decisions using intermediate execution observations.
\textbf{\textsc{Global}} indicates whether the model can generate a complete workflow before execution.
\textbf{Reference Workflow} indicates whether executable reference workflow structures are available for evaluation.
\textbf{State Grading} evaluates the resulting environment state.
\textbf{Side-Effect Check} evaluates external effects produced during execution.
\cmark~= yes, \pmark~= partial, and \xmark~= no.
}
\label{tab:benchmark_comparison}

\resizebox{\textwidth}{!}{
\begin{tabular}{lllcccccc}
\toprule

\textbf{Benchmark}
& \textbf{Domain}
& \textbf{Interface}
& \multicolumn{2}{c}{\textbf{Execution}}
& \multicolumn{2}{c}{\textbf{Global Orchestration}}
& \multicolumn{2}{c}{\textbf{Outcome Evaluation}} \\

\cmidrule(lr){4-5}
\cmidrule(lr){6-7}
\cmidrule(lr){8-9}

&
&
&
\textbf{E2E Exec.}
& \textbf{\textsc{Step-by-Step}}
& \textbf{\textsc{Global}}
& \textbf{Reference Workflow}
& \textbf{State Grading}
& \textbf{Side-Effect Check} \\

\midrule

WebArena~\citep{zhou2024webarena}
& Web / GUI & Browser
& \cmark & \cmark & \xmark & \xmark & \cmark & \pmark \\

WorkArena++~\citep{boisvert2024workarena++}
& Enterprise / GUI & Browser / GUI
& \cmark & \cmark & \xmark & \xmark & \cmark & \pmark \\

OSWorld~\citep{xie2024osworld}
& OS / GUI & Desktop GUI
& \cmark & \cmark & \xmark & \xmark & \pmark & \pmark \\

AppWorld~\citep{trivedi2024appworld}
& API / Tool & APIs
& \cmark & \cmark & \xmark & \xmark & \cmark & \cmark \\

$\tau$-bench~\citep{yao2024tau}
& API / Tool & Tools + User
& \cmark & \cmark & \xmark & \xmark & \cmark & \cmark \\

ToolBench~\citep{qin2024toolllm}
& API / Tool & APIs
& \pmark & \cmark & \xmark & \xmark & \pmark & \xmark \\

\midrule

AutomationBench~\citep{shepard2026automationbench}
& Workflow Agent & SaaS APIs
& \cmark & \cmark & \pmark & \pmark & \cmark & \cmark \\

WorFBench~\citep{qiao2025benchmarking}
& Workflow Generation & Workflow Graph
& \xmark & \xmark & \cmark & \cmark & \xmark & \xmark \\

WorkflowBench~\citep{fan2025workflowllm}
& Workflow Generation & APIs + Workflow
& \pmark & \pmark & \cmark & \cmark & \pmark & \xmark \\

T-Eval~\citep{DBLP:conf/acl/ChenDZLLZZZLCZ24}
& General Planning & Text / Tools
& \pmark & \pmark & \cmark & \pmark & \xmark & \xmark \\

\midrule

\textbf{\textsc{AgentOrch}}
& \textbf{Enterprise Workflow}
& \textbf{iPaaS Simulator}
& \cmark & \cmark & \cmark & \cmark & \cmark & \cmark \\

\bottomrule
\end{tabular}
}

\end{table*}

\vspace{-2mm}
\section{\textsc{AgentOrch}: Benchmarking Enterprise Workflow Orchestration}
\label{sec:benchmark}
\vspace{-2mm}
\textsc{AgentOrch} is designed to quantify the gap between global and step-by-step orchestration and to study the factors that determine whether this gap can be narrowed. As illustrated in Figure~\ref{fig:benchmark_pipeline}, we construct the benchmark from executable enterprise workflows and evaluate orchestration methods under matched task specifications, tool interfaces, runtime variants, execution environments, and evaluation criteria. Each workflow is evaluated under multiple runtime conditions, allowing us to compare how the same task is handled when orchestration is completed before execution versus through step-by-step interaction.

\textbf{Benchmark Overview.}
\textsc{AgentOrch} contains 650 executable standard operating procedures (SOPs) and 3,220 runtime variants. Each source recipe defines one SOP and serves as its reference workflow. For each SOP, we construct a natural-language task description (Appendix~\ref{app:bench_task_desc}) and multiple runtime variants. The task description and reference workflow remain fixed, while variants instantiate different inputs and execution conditions. This design allows the same workflow task to be evaluated under both \textsc{Global} and \textsc{Step-by-Step} orchestration.

\textbf{Task Specification.}
\label{sec:task-specification}
Each SOP has a fixed \textit{task specification} containing the task description, public input/output schemas, execution context, workspace resources, communication templates when applicable, and 64 available tool interfaces. Requirements are presented in non-causal order, requiring the model to infer tool calls, dependencies, control flow, and argument bindings. The specification excludes the reference workflow, reference traces, hidden environment state, fault settings, and concrete runtime inputs.

\subsection{Execution Environment and Runtime Variants}
\label{sec:execution-environment}

\textbf{Environment Construction.}
We map recipe operations to a fixed catalog of 64 tool interfaces spanning services such as Slack, Jira, Salesforce, Google Sheets, Google Drive, Snowflake, and Workato. We implement real and simulated backends behind the same interfaces: real connectors are used to collect reference behavior, while model-generated workflows are evaluated in simulation. The simulated backend reproduces the benchmark-relevant connector behavior while enabling isolated and reproducible evaluation without external API rate limits, quotas, or mutable service state.

\textbf{Runtime Variants.}
Each SOP is associated with multiple runtime variants that instantiate concrete inputs and execution conditions. These variants exercise different control-flow paths, loop cardinalities, and error conditions while preserving the same task description and reference workflow. Executions are isolated across variants. Construction details are provided in Appendix~\ref{app:bench_input_variant}.

\textbf{Reference Workflow and Ground Truth.}
For each SOP, the source recipe serves as the reference workflow. We execute it on each runtime variant using the real connector backend and record the execution trace, intermediate tool results, and terminal output. These executions define the reference behavior used for evaluation.

\subsection{Orchestration Protocols}
\label{subsec:orchestration-protocols}

We compare \textsc{Global} and \textsc{Step-by-Step} as the two primary orchestration protocols. We additionally study \textsc{Local} for incremental workflow construction and \textsc{Replan} for execution-feedback improvement.

\textbf{\textsc{Global}.}
The model generates one complete workflow before execution and before observing concrete runtime inputs. The workflow is then fixed and reused across all variants of the SOP without further LLM interaction.

\textbf{\textsc{Step-by-Step}.}
The protocol uses an action--observation loop related to ReAct~\citep{yao2022react}. Each variant is an independent episode. The model receives the task specification and concrete runtime input, produces one tool call, observes its result, and selects the next action until termination or the execution limit.

\textsc{Global} and \textsc{Step-by-Step} differ in the information available during execution: \textsc{Step-by-Step} receives the concrete runtime input and intermediate tool results, whereas \textsc{Global} constructs its workflow before observing either. We therefore use their performance difference as the gap between the two orchestration settings, rather than attributing it solely to execution feedback.

\textbf{\textsc{Local}.}
\textsc{Local} receives the same task specification as \textsc{Global} and observes neither concrete runtime inputs nor execution results, but constructs the workflow incrementally through an append-only editor. After each proposed edit, a validator either accepts the node and returns the updated workflow draft or rejects it with an error. Thus, \textsc{Local} tests whether incremental construction with validator feedback improves planning without runtime execution feedback.

\textbf{\textsc{Replan}.}
\textsc{Replan} studies feedback-driven self-improvement without model-weight updates, a principle also used by Reflexion~\citep{shinn2023reflexion}. Starting from the \textsc{Global} workflow, we execute it on runtime variants and provide the model with its workflow and execution feedback, including tool calls, arguments, results or errors, control-flow outcomes, and terminal returns. The model then generates a revised complete workflow. Reference workflows, reference executions, and evaluation scores are never provided.

\subsection{Evaluation Metrics}
\label{subsec:evaluation}

\textbf{Orchestration Accuracy.}
Our primary metric is the average of Execution Trace Accuracy and Side-Effect Accuracy for each runtime variant.
\textit{Execution Trace Accuracy.}
Execution Trace Accuracy measures agreement with the reference execution trace and averages five binary components: tool selection, tool order, tool execution, required arguments, and control path.
\textit{Side-Effect Accuracy.}
Side-Effect Accuracy measures whether execution produces the required state changes without missing, duplicated, or unintended effects.

\textbf{Final Accuracy.}
Final Accuracy measures whether both the required side effects and terminal output are correct. It is counted as correct only when both Side-Effect Accuracy and Terminal Output Accuracy are correct. Terminal Output Accuracy checks whether the terminal response matches the expected output when the SOP specifies one.

\textbf{Error Rate.}
Error Rate measures the fraction of runtime variants that do not produce a scoreable execution because of planning, model, infrastructure, or evaluation failures. These variants remain part of the evaluation and receive zero for the accuracy metrics rather than being excluded.

\subsection{Training for Global Orchestration}
\label{sec:training}

We next study whether global orchestration can be improved through targeted training. To align training with the \textsc{Global} setting, each training example requires the model to generate one complete workflow for an SOP without using any concrete runtime variant. We construct these examples from a separate set of 3,000 source recipes. The 3,000 training recipes are disjoint from the \textsc{AgentOrch} benchmark by source recipe ID.

\textbf{Training Data Construction}
\label{subsec:training-data}
For each source recipe, we construct a task specification following the same procedure used for \textsc{AgentOrch}. The source recipe serves as the reference workflow. This produces one training pair per recipe: the task specification is the model input, and the complete workflow is the target output. The target preserves the full workflow structure of the source recipe, including conditional branches, loops, exception handling, tool calls, and dependencies between tool results and later operations. Runtime variants are not used to specialize the target. The resulting training set contains 3,000 training pairs. Across the training set, the reference workflows use 1,033 tool operations spanning 305 namespaces.
\begin{wrapfigure}{r}{0.60\textwidth}
    \centering
    \includegraphics[width=\linewidth]{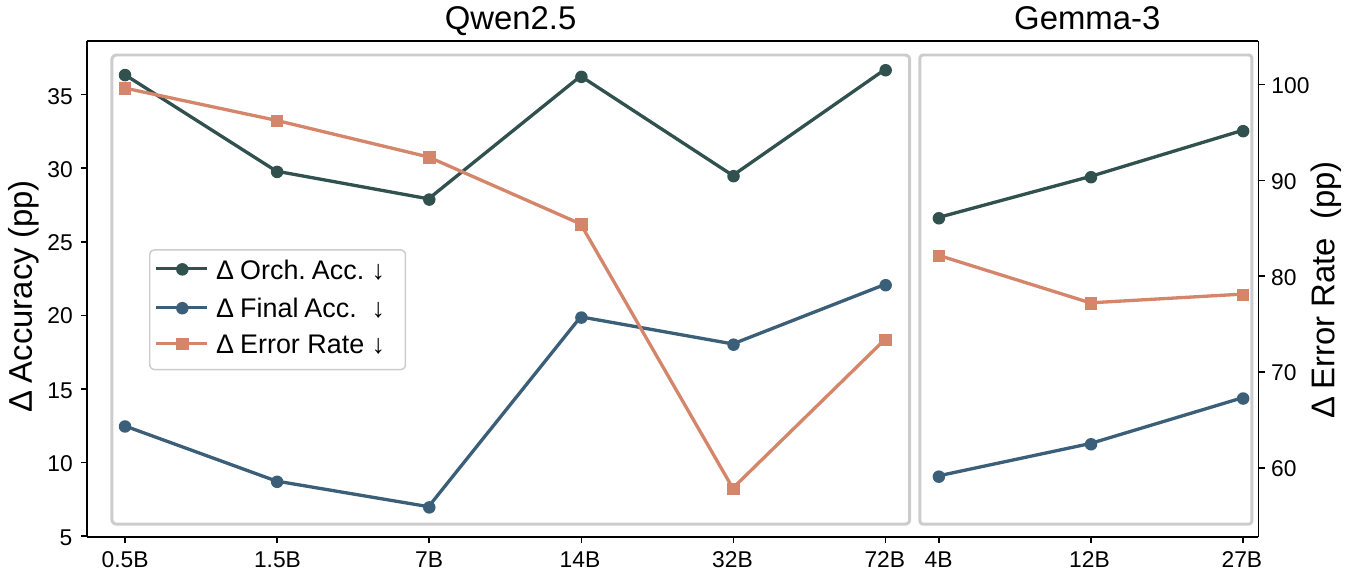}
    \caption{\small \textbf{Scaling does not consistently narrow the GLOBAL--STEP-BY-STEP gap.}
    Across Qwen2.5 and Gemma-3, the differences in Orchestration Accuracy, Final Accuracy, and Error Rate do not decrease monotonically with model size.}
    \vspace{-8mm}
\label{fig:rq1_scaling}
\end{wrapfigure}

\subsection{Training Methods}
\label{subsec:training-methods}
\textbf{Supervised Fine-Tuning.}
We first apply supervised fine-tuning to train the model to generate the reference workflow from the task specification. The training objective is applied to the generated workflow, while the task specification is provided as input context. We use the same training formulation across all evaluated base models.

\textbf{Reinforcement Learning.}
We further apply reinforcement learning to optimize generated workflows against the reference workflow. For each sampled workflow, the reward measures whether the output is valid and how closely it matches the reference workflow in tool calls, tool arguments, and workflow structure. The reward does not execute the generated workflow or use the simulated connector backend. We optimize the model using Group Relative Policy Optimization (GRPO)~\citep{shao2024deepseekmath}.
Training configurations for both methods are provided in Appendix~\ref{app:training_details}.

\section{Experiments}

We first characterize what does and does not explain the \textsc{Global}--\textsc{Step-by-Step} gap. We then study three routes for narrowing it, each drawing on a stronger signal than the last: incremental workflow construction without runtime execution feedback, workflow revision using execution feedback, and targeted training on reference workflows. Additional evaluation details are provided in Appendix~\ref{app:evaluation_details}.

\begin{mdframed}[userdefinedwidth=.99\linewidth,align=center,skipabove=3pt,skipbelow=3pt,innerleftmargin=6pt,innerbottommargin=6pt,innertopmargin=6pt,roundcorner=3pt,backgroundcolor=darkgreen!10,linecolor=gray]
\noindent \textbf{RQ1}: \textit{Do improvements in model capability translate into stronger global orchestration and a smaller gap to step-by-step interaction?}
\end{mdframed}
Improvements in model capability may raise both \textsc{Global} and \textsc{Step-by-Step} performance without reducing the difference between them. We therefore distinguish two outcomes: absolute improvement in \textsc{Global} performance and reduction in the \textsc{Global}--\textsc{Step-by-Step} gap. We examine three model-side factors: scale, model generation, and reasoning. Full results are provided in Appendix~\ref{app:rq1_full}.

\textbf{Model Scale Does Not Consistently Improve Global Orchestration or Narrow the Gap.}
Figure~\ref{fig:rq1_scaling} compares \textsc{Global} and \textsc{Step-by-Step} across model sizes within Qwen2.5 and Gemma-3. Neither \textsc{Global} Orchestration Accuracy nor the gap changes monotonically with scale. For Qwen2.5, \textsc{Global} Orchestration Accuracy drops from 16.18\% at 32B to 9.91\% at 72B, and the gap decreases from 36.33 pp at 0.5B to 27.87 pp at 7B before rising to 36.66 pp at 72B; for Gemma-3, the gap grows from 26.60 pp at 4B to 32.52 pp at 27B. \textsc{Step-by-Step} also retains higher Final Accuracy and lower Error Rate at every evaluated size.
\begin{wraptable}{r}{0.49\linewidth}
\vspace{-3mm}
\centering
\caption{\small
\textbf{Conditioning global orchestration on concrete runtime inputs does not narrow the gap.}
Orchestration Accuracy (\%); $\Delta$ is \textsc{Input-Conditioned Global} minus \textsc{Global} in percentage points.
}
\label{tab:ablation}
\scriptsize
\setlength{\tabcolsep}{3.2pt}
\renewcommand{\arraystretch}{1.04}

\resizebox{\linewidth}{!}{
\begin{tabular}{lrrrr}
\toprule
\textbf{Model}
& \textbf{\textsc{Global}}
& \textbf{\textsc{Input-Cond. Global}}
& \textbf{\textsc{Step-by-Step}}
& $\boldsymbol{\Delta}$ \\
\midrule
Qwen3-4B      & 13.75 & 13.04 & 41.84 & -0.71 \\
Gemma-4-12B   & 14.26 & 12.73 & 49.83 & -1.53 \\
Gemma-4-26B   & 4.05  & 3.66  & 52.10 & -0.39 \\
Claude Opus 5 & 46.44 & 48.79 & 63.52 & +2.35 \\
DeepSeek-V4   & 27.01 & 25.62 & 58.99 & -1.39 \\
Qwen3.8-27B   & 28.57 & 29.19 & 53.31 & +0.62 \\
\midrule
\textbf{Average}
& 22.35 & 22.18 & 53.27 & \textbf{-0.18} \\
\bottomrule
\end{tabular}
}

\vspace{-3mm}
\end{wraptable}

\textbf{Newer Generations Improve Global Orchestration in Most Comparisons, but Do Not Consistently Narrow the Gap.}
Across the generation pairs (Table~\ref{tab:rq1_generation}), newer generations improve \textsc{Global} Orchestration Accuracy in four pairs, by 5.80--17.98 pp, with only Llama decreasing slightly ($-0.36$ pp). However, the gap narrows only for Qwen (3.88 pp) and GPT (18.64 pp), and widens for Gemma, Llama, and GLM (1.39--6.18 pp). The Final Accuracy gap narrows only for GPT, while \textsc{Global} Error Rate decreases in all five pairs. Thus, newer generations generally strengthen \textsc{Global}, but stronger absolute performance does not consistently translate into a smaller gap.

\textbf{Reasoning Does Not Consistently Improve Global Orchestration or Narrow the Gap.}
Enabling reasoning increases \textsc{Global} Orchestration Accuracy for four of eight models and decreases it for the other four (Table~\ref{tab:rq1_reasoning}, Table~\ref{tab:rq1_reasoning_add}), with changes ranging from +13.44 pp for Claude Opus 5 to $-9.87$ pp for DeepSeek-V4-Flash. The corresponding gap narrows for only three models and widens for five. Thus, reasoning provides neither a consistent absolute benefit to \textsc{Global} nor a consistent reduction in the gap.

\begin{table}[t]
\centering
\caption{\small
\textbf{Newer model generations generally improve global orchestration,
but do not consistently narrow the gap to step-by-step interaction.}
Scores are reported in percent; Gap and $\Delta$ are reported in percentage points (pp).
For accuracy metrics, Gap is \textsc{Step-by-Step} minus \textsc{Global};
for Error Rate, Gap is \textsc{Global} minus \textsc{Step-by-Step}.
$\Delta$ denotes newer generation minus earlier generation.
}
\label{tab:rq1_generation}

\scriptsize
\setlength{\tabcolsep}{2.0pt}
\renewcommand{\arraystretch}{1.05}

\resizebox{\textwidth}{!}{%
\begin{tabular}{@{}ll
                ccc
                ccc
                ccc
                ccc
                ccc@{}}
\toprule

&&
\multicolumn{3}{c}{\textbf{Qwen}} &
\multicolumn{3}{c}{\textbf{Gemma}} &
\multicolumn{3}{c}{\textbf{Llama}} &
\multicolumn{3}{c}{\textbf{GPT}} &
\multicolumn{3}{c}{\textbf{GLM}}
\\

\cmidrule(lr){3-5}
\cmidrule(lr){6-8}
\cmidrule(lr){9-11}
\cmidrule(lr){12-14}
\cmidrule(lr){15-17}

\textbf{Metric}
& \textbf{Setting}

& \shortstack{Qwen2.5\\3B}
& \shortstack{Qwen3\\4B}
& $\Delta$

& \shortstack{Gemma-3\\12B}
& \shortstack{Gemma-4\\12B}
& $\Delta$

& \shortstack{Llama-3.1\\70B}
& \shortstack{Llama-3.3\\70B}
& $\Delta$

& GPT-3.5
& GPT-4o
& $\Delta$

& \shortstack{GLM-4.5\\FP8}
& \shortstack{GLM-5.2\\FP8}
& $\Delta$
\\

\midrule

\textbf{Orch. $\uparrow$}
& \textsc{Global}
& 2.19
& 13.75
& \multicolumn{1}{H}{+11.56}
& 8.46
& 14.26
& \multicolumn{1}{H}{+5.80}
& 10.81
& 10.45
& \multicolumn{1}{H}{-0.36}
& 10.11
& 28.09
& \multicolumn{1}{H}{+17.98}
& 12.93
& 19.23
& \multicolumn{1}{H}{+6.30}
\\

& \textsc{Step-by-Step}
& 34.16
& 41.84
& \multicolumn{1}{H}{+7.68}
& 37.85
& 49.83
& \multicolumn{1}{H}{+11.98}
& 38.92
& 42.98
& \multicolumn{1}{H}{+4.06}
& 50.78
& 50.12
& \multicolumn{1}{H}{-0.66}
& 54.47
& 62.16
& \multicolumn{1}{H}{+7.69}
\\

\cmidrule(lr){2-17}

& \textbf{Gap}
& 31.97
& 28.09
& \multicolumn{1}{H}{\textbf{-3.88}}
& 29.39
& 35.57
& \multicolumn{1}{H}{\textbf{+6.18}}
& 28.11
& 32.53
& \multicolumn{1}{H}{\textbf{+4.42}}
& 40.67
& 22.03
& \multicolumn{1}{H}{\textbf{-18.64}}
& 41.54
& 42.93
& \multicolumn{1}{H}{\textbf{+1.39}}
\\

\midrule

\textbf{Final $\uparrow$}
& \textsc{Global}
& 0.59
& 3.91
& \multicolumn{1}{H}{+3.32}
& 1.83
& 4.07
& \multicolumn{1}{H}{+2.24}
& 3.35
& 4.29
& \multicolumn{1}{H}{+0.94}
& 4.32
& 10.34
& \multicolumn{1}{H}{+6.02}
& 5.19
& 8.57
& \multicolumn{1}{H}{+3.38}
\\

& \textsc{Step-by-Step}
& 9.50
& 17.30
& \multicolumn{1}{H}{+7.80}
& 13.07
& 27.76
& \multicolumn{1}{H}{+14.69}
& 14.41
& 20.31
& \multicolumn{1}{H}{+5.90}
& 30.06
& 29.13
& \multicolumn{1}{H}{-0.93}
& 38.29
& 48.20
& \multicolumn{1}{H}{+9.91}
\\

\cmidrule(lr){2-17}

& \textbf{Gap}
& 8.91
& 13.39
& \multicolumn{1}{H}{\textbf{+4.48}}
& 11.24
& 23.69
& \multicolumn{1}{H}{\textbf{+12.45}}
& 11.06
& 16.02
& \multicolumn{1}{H}{\textbf{+4.96}}
& 25.74
& 18.79
& \multicolumn{1}{H}{\textbf{-6.95}}
& 33.10
& 39.63
& \multicolumn{1}{H}{\textbf{+6.53}}
\\

\midrule

\textbf{Error $\downarrow$}
& \textsc{Global}
& 94.78
& 61.96
& \multicolumn{1}{H}{-32.82}
& 77.17
& 65.78
& \multicolumn{1}{H}{-11.39}
& 72.24
& 71.46
& \multicolumn{1}{H}{-0.78}
& 75.09
& 35.40
& \multicolumn{1}{H}{-39.69}
& 75.68
& 61.34
& \multicolumn{1}{H}{-14.34}
\\

& \textsc{Step-by-Step}
& 0.50
& 0.06
& \multicolumn{1}{H}{-0.44}
& 0.03
& 0.16
& \multicolumn{1}{H}{+0.13}
& 0.03
& 0.00
& \multicolumn{1}{H}{-0.03}
& 0.19
& 0.16
& \multicolumn{1}{H}{-0.03}
& 0.09
& 0.12
& \multicolumn{1}{H}{+0.03}
\\

\cmidrule(lr){2-17}

& \textbf{Gap}
& 94.28
& 61.90
& \multicolumn{1}{H}{\textbf{-32.38}}
& 77.14
& 65.62
& \multicolumn{1}{H}{\textbf{-11.52}}
& 72.21
& 71.46
& \multicolumn{1}{H}{\textbf{-0.75}}
& 74.90
& 35.24
& \multicolumn{1}{H}{\textbf{-39.66}}
& 75.59
& 61.22
& \multicolumn{1}{H}{\textbf{-14.37}}
\\

\bottomrule
\end{tabular}%
}

\end{table}

\begin{table*}[t]
\centering
\caption{\small \textbf{Reasoning does not consistently improve global orchestration.}
We report global-mode performance for base and thinking variants across models. For each base--thinking pair, the better-performing score is highlighted and bolded (higher is better for accuracy metrics; lower is better for error rate).}
\label{tab:rq1_reasoning}

\scriptsize
\setlength{\tabcolsep}{2.2pt}
\renewcommand{\arraystretch}{1.05}

\resizebox{\textwidth}{!}{%
\begin{tabular}{@{}l
                rr@{\hspace{2pt}\vrule\hspace{2pt}}
                rr@{\hspace{2pt}\vrule\hspace{2pt}}
                rr@{\hspace{2pt}\vrule\hspace{2pt}}
                rr@{\hspace{2pt}\vrule\hspace{2pt}}
                rr@{\hspace{2pt}\vrule\hspace{2pt}}
                rr@{\hspace{2pt}\vrule\hspace{2pt}}
                rr@{\hspace{2pt}\vrule\hspace{2pt}}
                rr@{}}

\toprule

& \multicolumn{2}{c}{\textbf{Qwen3-4B}}
& \multicolumn{2}{c}{\textbf{Qwen3-30B}}
& \multicolumn{2}{c}{\textbf{Qwen3.8-27B}}
& \multicolumn{2}{c}{\textbf{DeepSeek-V4}}
& \multicolumn{2}{c}{\textbf{GLM-4.5-Air}}
& \multicolumn{2}{c}{\textbf{GLM-4.5-FP8}}
& \multicolumn{2}{c}{\textbf{GLM-5.2-FP8}}
& \multicolumn{2}{c}{\textbf{Claude-Opus-5}}
\\

\cmidrule(lr){2-3}
\cmidrule(lr){4-5}
\cmidrule(lr){6-7}
\cmidrule(lr){8-9}
\cmidrule(lr){10-11}
\cmidrule(lr){12-13}
\cmidrule(lr){14-15}
\cmidrule(lr){16-17}

\textbf{Metric}
& Base & Think
& Base & Think
& Base & Think
& Base & Think
& Base & Think
& Base & Think
& Base & Think
& Base & Think
\\

\midrule

Orch.
& \best{13.75\%} & 10.25\%
& 8.35\%  & \best{14.75\%}
& 28.57\% & \best{33.02\%}
& \best{27.01\%} & 17.14\%
& \best{12.28\%} & 10.72\%
& 12.93\% & \best{20.09\%}
& \best{19.23\%} & 17.75\%
& 46.44\% & \best{59.88\%}
\\

Final
& \best{3.91\%} & 2.70\%
& 2.95\% & \best{5.19\%}
& 12.67\% & \best{17.05\%}
& \best{11.43\%} & 9.66\%
& \best{4.78\%} & 4.25\%
& 5.19\% & \best{8.48\%}
& 8.57\% & \best{9.29\%}
& 23.54\% & \best{32.33\%}
\\

Side
& \best{16.40\%} & 13.04\%
& 9.44\% & \best{16.83\%}
& 31.80\% & \best{35.71\%}
& \best{29.94\%} & 17.52\%
& \best{14.22\%} & 11.61\%
& 13.88\% & \best{21.46\%}
& \best{21.15\%} & 18.66\%
& 50.16\% & \best{63.98\%}
\\

Trace
& \best{11.11\%} & 7.45\%
& 7.26\% & \best{12.68\%}
& 25.33\% & \best{30.33\%}
& \best{24.07\%} & 16.76\%
& \best{10.34\%} & 9.83\%
& 11.99\% & \best{18.72\%}
& \best{17.31\%} & 16.84\%
& 42.72\% & \best{55.78\%}
\\

Term.
& \best{16.30\%} & 8.42\%
& 13.73\% & \best{20.12\%}
& 28.98\% & \best{35.16\%}
& \best{28.98\%} & 20.06\%
& 11.71\% & \best{13.45\%}
& 10.81\% & \best{19.88\%}
& \best{18.94\%} & 14.32\%
& 42.24\% & \best{48.29\%}
\\

\midrule

\textbf{Mean Acc. $\uparrow$}
& \best{12.29\%} & 8.37\%
& 8.35\% & \best{13.91\%}
& 25.47\% & \best{30.25\%}
& \best{24.29\%} & 16.23\%
& \best{10.67\%} & 9.97\%
& 10.96\% & \best{17.73\%}
& \best{17.04\%} & 15.37\%
& 41.02\% & \best{52.05\%}
\\

\textbf{Error $\downarrow$}
& \best{61.96\%} & 77.76\%
& 76.43\% & \best{60.81\%}
& 43.88\% & \best{39.47\%}
& \best{43.51\%} & 67.67\%
& \best{73.76\%} & 74.63\%
& 75.68\% & \best{58.35\%}
& \best{61.34\%} & 73.01\%
& 19.16\% & \best{6.21\%}
\\

\bottomrule

\end{tabular}%
}

\end{table*}

\textbf{Ablation: Access to Runtime Inputs Alone Does Not Explain the Gap}
We next test whether access to concrete runtime inputs accounts for the protocol gap. \textsc{Input-Conditioned Global} receives the runtime input but generates the complete workflow before execution, without intermediate execution results. Across six models, this changes Orchestration Accuracy by $-0.18$ pp on average; the mean gap to \textsc{Step-by-Step} changes from 30.92 pp to 31.09 pp (Table~\ref{tab:ablation}). Providing the runtime input alone therefore leaves the average gap largely unchanged. The comparison still jointly varies sequential decision making and intermediate execution information, so it does not isolate their contributions.

\par
\clearwrap

\begin{mdframed}[userdefinedwidth=.99\linewidth,align=center,skipabove=3pt,skipbelow=3pt,innerleftmargin=6pt,innerbottommargin=6pt,innertopmargin=6pt,roundcorner=3pt,backgroundcolor=darkgreen!10,linecolor=gray]
\noindent \textbf{RQ2}: \textit{Can incremental workflow construction improve global orchestration without runtime execution feedback?}
\end{mdframed}

\begin{wraptable}{r}{0.48\linewidth}
\vspace{-6mm}
\centering
\caption{\small
\textbf{Incremental plan construction can improve global orchestration without execution feedback.}
Orchestration Accuracy (\%); $\Delta$ is \textsc{Local} minus \textsc{Global}. 
}
\label{tab:local_view}
\small
\setlength{\tabcolsep}{5pt}
\renewcommand{\arraystretch}{1.05}

\resizebox{\linewidth}{!}{
\begin{tabular}{lrrr}
\toprule
\textbf{Model}
& \textbf{\textsc{Global}}
& \textbf{\textsc{Local}}
& $\boldsymbol{\Delta}$ \\
\midrule
Llama-3.3-70B          & 10.4 & 34.9 & \textbf{+24.5} \\
Qwen2.5-3B             & 2.2  & 22.8 & \textbf{+20.6} \\
Mistral-Small-3.1-24B  & 14.1 & 27.3 & \textbf{+13.1} \\
Qwen3.8-27B-Thinking   & 33.0 & 41.5 & \textbf{+8.5} \\
Qwen3-30B-A3B          & 8.4  & 8.5  & \textbf{+0.1} \\
Qwen3-4B               & 13.8 & 6.9  & -6.9 \\
\bottomrule
\end{tabular}
}

\end{wraptable}

\textbf{Incremental Workflow Construction Improves Global Orchestration for Most Evaluated Models.}
Since stronger models alone do not reliably narrow the gap, we next change how the workflow is constructed while keeping the available information fixed. \textsc{Local} and \textsc{Global} receive the same task specification and observe neither concrete runtime inputs nor execution results; they differ only in whether the workflow is generated in one shot or constructed incrementally with validation after each edit (Section~\ref{subsec:orchestration-protocols}). \textsc{Local} improves Orchestration Accuracy by 9.98 pp on average (Table~\ref{tab:local_view}), with the largest gains for Llama-3.3-70B (+24.5 pp) and Qwen2.5-3B (+20.6 pp). Qwen3-4B is the only model that declines ($-6.9$ pp).

\par
\clearwrap

\begin{mdframed}[userdefinedwidth=.99\linewidth,align=center,skipabove=3pt,skipbelow=3pt,innerleftmargin=6pt,innerbottommargin=6pt,innertopmargin=6pt,roundcorner=3pt,backgroundcolor=darkgreen!10,linecolor=gray]
\noindent \textbf{RQ3}: \textit{Can execution feedback improve global workflows, and where do the gains come from?}
\end{mdframed}

\begin{figure}[htbp]
    \centering
    \includegraphics[width=\textwidth]{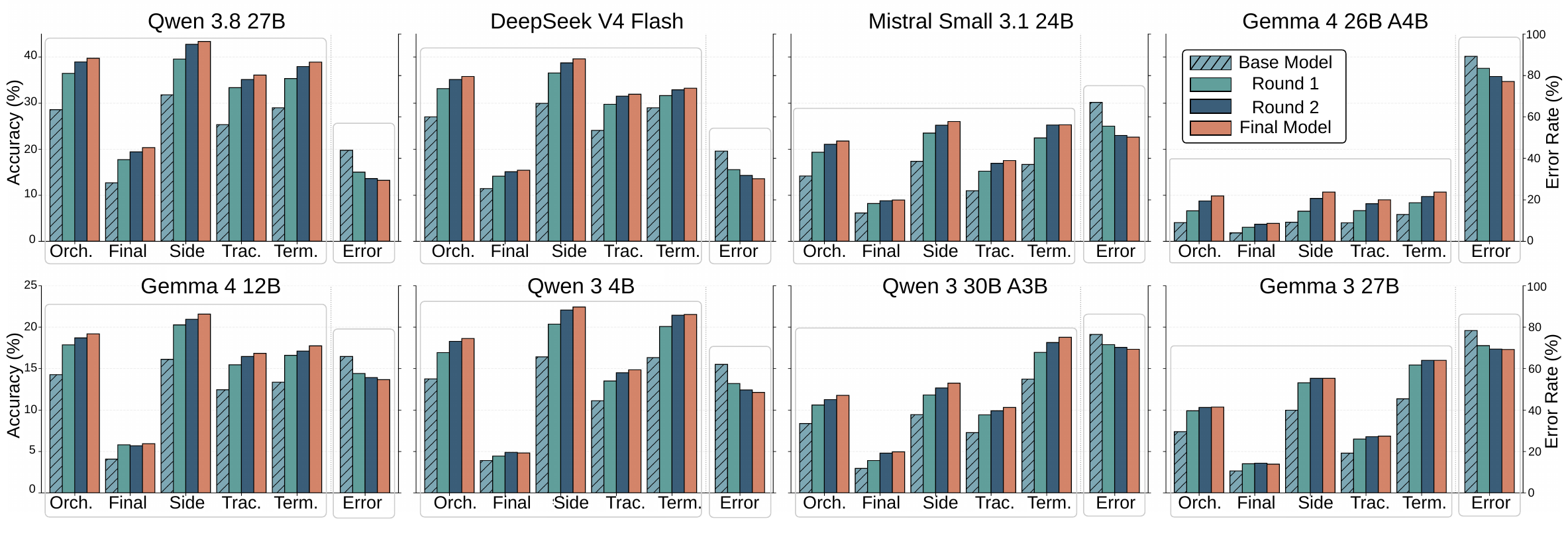}
    \vspace{-2mm}
    \caption{\small \textbf{\textsc{Replan} consistently improves global orchestration across models.}
Performance over Rounds 0--3 for eight models when execution feedback is collected from all runtime variants of each SOP. Accuracy metrics generally increase across revision rounds, while Error Rate decreases.}
    \label{fig:rq2a}
    \vspace{-2mm}
\end{figure}

\textbf{Execution-Feedback Revision Improves Global Orchestration.}
RQ2 improves \textsc{Global} without observing any execution. We next allow the model to observe how its workflow executes. \textsc{Replan} executes the \textsc{Global} workflow, returns the execution feedback to the model, and asks it to revise the complete workflow. The revised workflow still executes without further LLM calls. Over three rounds, \textsc{Replan} improves Orchestration Accuracy for all eight evaluated models, by 6.18 pp on average, and reduces Error Rate for every model (Figure~\ref{fig:rq2a}).

\textbf{The Gains Transfer to Unseen Runtime Variants.}
We next examine whether the improvement is limited to the runtime variants used to construct the feedback. We provide feedback from only two variants of each SOP and evaluate on the remaining held-out variants. The average held-out gain is 6.13 pp, nearly identical to the 6.18 pp obtained with feedback from all variants (Table~\ref{tab:replan_heldout_variants}), and far more individual held-out variants improve than regress (Appendix~\ref{app:rq3_additional}). \textsc{Replan} therefore improves the workflow itself, not only its behavior on the feedback executions.
\begin{wraptable}{r}{0.49\linewidth}
\vspace{-3mm}
\centering
\caption{\small
\textbf{Decomposition of \textsc{Replan} gains.}
Repair and Valid: gains from initially invalid and initially valid workflows. Validation-only: revision using pre-execution validation errors without execution. Recovery $=\Delta/(\mathrm{Repair}+\mathrm{Valid})$.
}
\label{tab:rq3_repair}
\scriptsize
\setlength{\tabcolsep}{2.2pt}
\renewcommand{\arraystretch}{1.04}

\resizebox{\linewidth}{!}{
\begin{tabular}{lrrrrrr}
\toprule
& \multicolumn{3}{c}{\textbf{\textsc{Replan}}}
& \multicolumn{3}{c}{\textbf{Validation-only}} \\
\cmidrule(lr){2-4}
\cmidrule(lr){5-7}
\textbf{Model}
& \textbf{Repair}
& \textbf{Valid}
& \textbf{Share}
& $\boldsymbol{\Delta}$
& \textbf{Recovery}
& \textbf{R--S} \\
\midrule
Mistral-3.1-24B
& +7.09 & +0.52 & 93.2\%
& +6.10 & 80.2\% & +1.51 \\

Qwen3-30B-A3B
& +2.75 & +0.66 & 80.6\%
& +1.70 & 49.9\% & +1.71 \\

Qwen3-4B
& +4.88 & 0.00 & 100.0\%
& +4.20 & 86.1\% & +0.68 \\
\midrule
\textbf{Across models}
& \textbf{+4.91}
& \textbf{+0.39}
& \textbf{92.6\%}
& \textbf{+4.00}
& \textbf{75.5\%}
& \textbf{+1.30} \\
\bottomrule
\end{tabular}
}

\vspace{-4mm}
\end{wraptable}

\textbf{Most Gains Come from Fixing How Tools Are Called.}
Across the three models in the decomposition, revisions that turn invalid Round-0 workflows into valid ones account for 92.6\% of the aggregate \textsc{Replan} gain (Table~\ref{tab:rq3_repair}). In the eight-model analysis, most repaired workflows initially parse successfully but fail schema, type, or reference checks governing tool arguments and references between steps (Appendix~\ref{app:rq3_additional}). \textsc{Global} must specify these arguments and references before execution, while \textsc{Step-by-Step} can use observed results to construct subsequent calls. All initial failures in the repaired workflows are detectable before execution. Revising the same Round-0 workflows using only pre-execution validation errors recovers 75.5\% of the aggregate \textsc{Replan} gain across the three models. 

Together, these results show that global orchestration requires not only selecting the right tools, but also specifying before execution how each tool is called and how data flows between calls. Getting these specifications right is a major bottleneck of global orchestration.

\begin{table*}[t]
\centering
\caption{\small
\textbf{\textsc{Replan} improvements generalize to unseen runtime variants.}
We report performance before \textsc{Replan} (\textbf{Base}, Round 0) and after three \textsc{Replan} rounds (\textbf{Final}, Round 3).
At each round, the model receives execution feedback from only the two random runtime variants of each SOP, while evaluation is performed on the remaining held-out variants that are never used for feedback.
For Mean Accuracy and Error Rate, we additionally report the percentage-point change ($\Delta$).
Mean Accuracy averages Orchestration, Final, Side-effect, Execution-trace, and Terminal-output Accuracy.
Higher accuracy and lower Error Rate are better.
No model weights are updated.
}
\label{tab:replan_heldout_variants}
\scriptsize
\setlength{\tabcolsep}{2.6pt}
\renewcommand{\arraystretch}{1.08}

\resizebox{\textwidth}{!}{
\begin{tabular}{l*{16}{r}}
\toprule

\textbf{Model}
& \multicolumn{2}{c}{\textbf{Orch.} $\uparrow$}
& \multicolumn{2}{c}{\textbf{Final} $\uparrow$}
& \multicolumn{2}{c}{\textbf{Side} $\uparrow$}
& \multicolumn{2}{c}{\textbf{Trace} $\uparrow$}
& \multicolumn{2}{c}{\textbf{Term.} $\uparrow$}
& \multicolumn{3}{c}{\textbf{Mean Acc.} $\uparrow$}
& \multicolumn{3}{c}{\textbf{Error} $\downarrow$}
\\

\cmidrule(lr){2-3}
\cmidrule(lr){4-5}
\cmidrule(lr){6-7}
\cmidrule(lr){8-9}
\cmidrule(lr){10-11}
\cmidrule(lr){12-14}
\cmidrule(lr){15-17}

& \textbf{Base} & \textbf{Final}
& \textbf{Base} & \textbf{Final}
& \textbf{Base} & \textbf{Final}
& \textbf{Base} & \textbf{Final}
& \textbf{Base} & \textbf{Final}
& \textbf{Base} & \textbf{Final} & $\boldsymbol{\Delta}$
& \textbf{Base} & \textbf{Final} & $\boldsymbol{\Delta}$
\\

\midrule

Qwen3.8-27B
& 29.11 & 40.07
& 14.37 & 22.22
& 33.02 & 44.34
& 25.19 & 35.79
& 28.73 & 38.81
& 26.08 & 36.25 & \textbf{+10.16}
& 45.22 & 31.58 & \textbf{-13.64}
\\

DeepSeek-V4-Flash
& 27.21 & 35.46
& 12.35 & 16.69
& 30.90 & 40.62
& 23.51 & 30.29
& 28.48 & 33.18
& 24.49 & 31.25 & \textbf{+6.76}
& 44.86 & 31.27 & \textbf{-13.59}
\\

Mistral-Small-3.1-24B
& 14.52 & 21.89
& 6.93 & 9.82
& 18.29 & 26.72
& 10.74 & 17.06
& 16.38 & 24.55
& 13.37 & 20.01 & \textbf{+6.64}
& 67.60 & 51.21 & \textbf{-16.39}
\\

Gemma-4-26B-A4B
& 3.82 & 9.62
& 1.91 & 4.29
& 4.03 & 10.65
& 3.61 & 8.59
& 4.96 & 9.92
& 3.67 & 8.61 & \textbf{+4.95}
& 90.65 & 79.59 & \textbf{-11.06}
\\

Gemma-4-12B
& 14.47 & 20.24
& 4.34 & 6.67
& 16.90 & 23.98
& 12.04 & 16.51
& 12.87 & 17.57
& 12.12 & 16.99 & \textbf{+4.87}
& 66.93 & 54.68 & \textbf{-12.25}
\\

Qwen3-4B
& 14.24 & 19.13
& 4.81 & 6.61
& 17.31 & 23.62
& 11.17 & 14.65
& 15.71 & 21.55
& 12.65 & 17.11 & \textbf{+4.46}
& 63.62 & 50.49 & \textbf{-13.13}
\\

Qwen3-30B-A3B
& 8.50 & 11.14
& 3.31 & 5.27
& 10.03 & 12.66
& 6.98 & 9.61
& 12.92 & 17.98
& 8.35 & 11.33 & \textbf{+2.98}
& 77.26 & 71.11 & \textbf{-6.15}
\\

Gemma-3-27B
& 7.80 & 11.13
& 3.15 & 4.39
& 10.90 & 15.30
& 4.70 & 6.96
& 10.96 & 15.50
& 7.50 & 10.66 & \textbf{+3.15}
& 78.55 & 69.15 & \textbf{-9.40}
\\

\bottomrule
\end{tabular}
}
\end{table*}

\begin{mdframed}[userdefinedwidth=.99\linewidth,align=center,skipabove=3pt,skipbelow=3pt,innerleftmargin=6pt,innerbottommargin=6pt,innertopmargin=6pt,roundcorner=3pt,backgroundcolor=darkgreen!10,linecolor=gray]
\noindent \textbf{RQ4}: \textit{Can stronger global orchestration be learned through targeted training?}
\end{mdframed}

\textbf{Training on Complete Workflows Improves Global Orchestration for All Models.}
RQ2 and RQ3 improve \textsc{Global} at test time, and RQ3 shows that its main obstacle is producing a workflow with correct arguments and references. We next ask whether training can teach models to produce such workflows directly. We train five models on 3,000 source recipes disjoint from \textsc{AgentOrch}, each pairing a task specification with a complete reference workflow (Section~\ref{sec:training}). Supervised fine-tuning (SFT) improves Orchestration Accuracy and Final Accuracy for all five models, and taking the better of SFT and RL for each model yields an average Orchestration Accuracy gain of 12.18 pp (Table~\ref{tab:rq3_training}). Error Rate drops even more sharply, by 13.02--40.99 pp; Qwen3.8-27B falls from 43.88\% to 17.08\%, below the 19.16\% of Claude Opus 5 under \textsc{Global}.

\textbf{Reinforcement Learning Adds Gains for Some Models.}
We next examine whether reinforcement learning (RL) further improves the SFT models. As shown in Table~\ref{tab:rq3_training}, RL further increases Orchestration Accuracy for three of five models: Qwen3-4B by 1.70 pp, Gemma-4-26B-A4B by 4.04 pp, and Gemma-4-12B by 3.74 pp. Error Rate also decreases for four models, including reductions of 8.38 pp for Gemma-4-26B-A4B and 11.18 pp for Gemma-4-12B. However, RL does not improve Orchestration Accuracy for Qwen3.8-27B or Gemma-4-31B, and Final Accuracy improves for only two models. Overall, RL can provide additional gains after SFT, but the benefit is not consistent across models or metrics.

\begin{table*}[t]
\centering
\caption{
\textbf{Targeted training improves global orchestration.}
Results under \textsc{Global}; $\Delta$ compares the better of SFT and RL with Base (pp).
}
\label{tab:rq3_training}
\scriptsize
\setlength{\tabcolsep}{2.2pt}
\renewcommand{\arraystretch}{1.08}

\resizebox{\textwidth}{!}{
\begin{tabular}{@{}lcccccccccccccccc@{}}
\toprule
& \multicolumn{4}{c}{\textbf{Orch. Acc. (\%) $\uparrow$}}
& \multicolumn{4}{c}{\textbf{Final Acc. (\%) $\uparrow$}}
& \multicolumn{4}{c}{\textbf{Trace Acc. (\%) $\uparrow$}}
& \multicolumn{4}{c}{\textbf{Error Rate (\%) $\downarrow$}} \\
\cmidrule(lr){2-5}
\cmidrule(lr){6-9}
\cmidrule(lr){10-13}
\cmidrule(l){14-17}

\textbf{Model}
& \textbf{Base} & \textbf{SFT} & \textbf{RL} & $\boldsymbol{\Delta}$
& \textbf{Base} & \textbf{SFT} & \textbf{RL} & $\boldsymbol{\Delta}$
& \textbf{Base} & \textbf{SFT} & \textbf{RL} & $\boldsymbol{\Delta}$
& \textbf{Base} & \textbf{SFT} & \textbf{RL} & $\boldsymbol{\Delta}$ \\
\midrule

Qwen3.8-27B
& 28.57 & \textbf{41.96} & 41.14 & \textbf{+13.39}
& 12.67 & \textbf{18.48} & 18.26 & \textbf{+5.81}
& 25.33 & \textbf{35.87} & 35.24 & \textbf{+10.54}
& 43.88 & \textbf{17.08} & 19.38 & \textbf{$-26.80$} \\

Qwen3-4B
& 13.75 & 25.46 & \textbf{27.16} & \textbf{+13.41}
& 3.91 & \textbf{10.78} & 10.71 & \textbf{+6.87}
& 11.11 & 19.80 & \textbf{21.43} & \textbf{+10.32}
& 61.96 & 40.99 & \textbf{36.12} & \textbf{$-25.84$} \\

Gemma-4-26B-A4B
& 4.05 & 19.89 & \textbf{23.93} & \textbf{+19.88}
& 1.80 & 8.45 & \textbf{9.35} & \textbf{+7.55}
& 3.99 & 15.40 & \textbf{18.66} & \textbf{+14.67}
& 89.22 & 56.61 & \textbf{48.23} & \textbf{$-40.99$} \\

Gemma-4-31B
& 12.71 & \textbf{19.55} & 19.10 & \textbf{+6.84}
& 4.78 & \textbf{9.78} & 9.22 & \textbf{+5.00}
& 12.07 & \textbf{16.03} & 15.93 & \textbf{+3.96}
& 74.94 & 57.17 & \textbf{56.55} & \textbf{$-18.39$} \\

Gemma-4-12B
& 14.26 & 17.92 & \textbf{21.66} & \textbf{+7.40}
& 4.07 & 8.51 & \textbf{8.54} & \textbf{+4.47}
& 12.44 & 14.29 & \textbf{17.33} & \textbf{+4.89}
& 65.78 & 63.94 & \textbf{52.76} & \textbf{$-13.02$} \\

\bottomrule
\end{tabular}
}
\end{table*}

\section{Conclusion}
We introduced \textsc{AgentOrch} to study how close global orchestration can come to step-by-step interaction on the same enterprise workflows. \textsc{Step-by-Step} remains more accurate for every evaluated model. The gap has shrunk by about 79\% from early models to the strongest recent model, but this narrowing is concentrated in the latest generation: neither model scale, newer generations, nor reasoning alone consistently narrows it. We identified three routes that narrow the gap further, each drawing on a stronger signal than the last: incremental workflow construction without execution feedback, revision with execution feedback that transfers to unseen runtime variants, and targeted training on reference workflows that improves every trained model. Our analysis further shows that global orchestration requires not only selecting the right tools, but also specifying before execution how each tool is called and how data flows between calls; getting these specifications right is a key lever for future work. The gap has not closed, but it is closing: global orchestration offers a promising path toward enterprise agents that approach step-by-step accuracy with fewer LLM calls.

\titlespacing*{\section}{0pt}{*1}{*1}
\titlespacing*{\subsection}{0pt}{*1.25}{*1.25}
\titlespacing*{\subsubsection}{0pt}{*1.5}{*1.5}

\setlength{\abovedisplayskip}{\baselineskip} %
\setlength{\abovedisplayshortskip}{0.5\baselineskip} %
\setlength{\belowdisplayskip}{\baselineskip}
\setlength{\belowdisplayshortskip}{0.5\baselineskip}

\clearpage
\appendix
\label{sec:append}
\part*{Appendix}
{
\setlength{\parskip}{-0em}
\startcontents[sections]
\printcontents[sections]{ }{1}{}
}

\setlength{\parskip}{.5em}
\section{Benchmark and Data Construction}
\subsection{Source Recipes and Filtering}
\label{app:bench_source}

We start from a sanitized collection of 803 executable Workato workspace recipes. We exclude 153 recipes with fewer than two reachable enabled actions, resulting in the 650 SOPs used in \textsc{AgentOrch}. Each retained source recipe defines one SOP and serves as its reference workflow.

The benchmark split is defined at the source-recipe level. Each recipe appears at most once in the benchmark, while multiple runtime variants may be associated with the same recipe. The resulting benchmark contains 650 SOPs and 3,220 runtime variants.

\subsection{Execution Environment Construction}
\label{app:bench_environment}

We construct the execution environment from the software operations used by the source recipes. These operations are mapped to the fixed set of 64 tool interfaces exposed by \textsc{AgentOrch}. The same tool names and argument interfaces are used across the benchmark.

\textbf{Real connector backend.}
The real connector backend invokes the corresponding external software services where supported. During benchmark construction, we execute each reference workflow using the real connector backend under its runtime variants. We record the active execution trace, intermediate tool results, and terminal output produced by these executions. These records are used to construct the corresponding reference execution data.

\textbf{Simulated connector backend.}
For scalable model evaluation, we implement a local simulated backend for the connector operations exercised by AGENTORCH. The simulated backend exposes the same tool names, argument schemas, and return interfaces as the corresponding real connectors, and reproduces the benchmark-relevant execution semantics required by the reference workflows, including state transitions, observable side effects, and supported error behavior.

The simulated backend is used as an evaluation substitute for the real connectors rather than as a different task environment. Evaluating every model-generated workflow directly against live services would be impractical at benchmark scale because external connectors impose API rate limits and quotas, incur service-dependent latency, and operate over mutable external state. The simulated backend instead provides deterministic, isolated, and reproducible executions while preserving the connector behavior exercised by the benchmark.

Simulation is also necessary for safety: executing model-generated workflows against live enterprise services would send real messages, modify real records, and produce side effects that cannot be undone.

Each runtime variant initializes a separate simulated connector state. State changes produced by one tool call are visible to subsequent calls within the same execution, and the environment is reset between executions so that different models, SOPs, and runtime variants do not interfere with one another.

All model evaluations reported in this paper execute model-generated workflows using the simulated connector backend. The real connector backend is used during benchmark construction to collect reference execution results rather than to execute the evaluated model.

\subsection{Task Description Generation}
\label{app:bench_task_desc}

Following WorkflowLLM~\cite{fan2025workflowllm}, we use a bottom-up workflow-to-description procedure to construct the natural-language task description for each SOP. Let $A_r$ denote the structured representation of reference workflow $r$, and let $D_r$ denote the associated connector documentation and tool schemas. The construction follows
\[
A_r + D_r
\xrightarrow{\quad} C_r
\xrightarrow{\quad} P_r
\xrightarrow{\quad} \{Q_r^{(k)}\}_{k=1}^{K_Q}
\xrightarrow{\quad} Q_r,
\]
where $C_r$ contains action-level semantic annotations, $P_r$ contains the workflow requirements across its control-flow paths, $\{Q_r^{(k)}\}$ contains candidate task descriptions, and $Q_r$ is the selected task description.

\begin{enumerate}[leftmargin=*]

    \item \textbf{Workflow representation.}
    We deterministically transcribe reference workflow $r$ into a structured representation $A_r$. The representation retains the executable actions and their arguments, literal values, formula expressions, branch conditions, loops, try/catch blocks, and terminal behavior. Actions without a documented tool interface are identified during this stage.

    \item \textbf{Action-level annotation.}
    A language model produces an annotation for each workflow node using $A_r$ and $D_r$. The annotation describes the corresponding operation, its arguments, dependencies on preceding results, branch conditions, and error-handling behavior. Each annotation is required to be grounded in information contained in $A_r$ or $D_r$.

    \item \textbf{Task-plan construction.}
    The action-level annotations are combined into a task plan $P_r$ covering the requirements represented by the workflow. The plan includes the public inputs, branch conditions, loop behavior, fallback behavior, and exception handling required by the reference workflow. Runtime-variant values and hidden connector state are not included.

    \item \textbf{Candidate task-description generation.}
    Using $A_r$, $C_r$, and $P_r$, the model generates $K_Q=3$ candidate task descriptions. Each candidate describes the workflow objective, required inputs, conditions, and expected behavior without exposing recipe node identifiers or the reference execution trace.

    The requirements are not presented in the execution order of the reference workflow, so their textual presentation does not reveal the solution sequence. However, ordering requirements prescribed by the source SOP, including dependencies and temporal constraints between operations, are retained in the task description. Thus, the model must infer the prescribed execution order from the stated workflow requirements rather than from their presentation order.

    \item \textbf{Consistency checking and selection.}
    We extract a semantic ledger $L_r$ from $A_r$ containing the public inputs, operational requirements, and terminal behavior represented by the workflow. A model-based critic compares each candidate task description with $L_r$ for two forms of consistency: whether required information from the workflow is represented in the description, and whether the description introduces requirements not supported by the workflow. The selected candidate becomes $Q_r$.

    \item \textbf{Variant invariance and hidden-information filtering.}
    Each SOP uses one task description $Q_r$ across all of its runtime variants. The description does not contain concrete variant inputs, hidden connector state, fault-injection settings, reference execution traces, or expected tool-call sequences.
\end{enumerate}

The selected task description $Q_r$ is included in the task specification described in Section~\ref{sec:task-specification}. The remaining components of the task specification, including schemas, workspace resources, communication templates, and tool interfaces, are supplied separately. Runtime information is provided according to the orchestration protocol described in Section~\ref{subsec:orchestration-protocols}; in particular, \textsc{Global} does not receive the concrete runtime input when generating its workflow.

\subsection{Runtime Variant Generation}
\label{app:bench_input_variant}

Each SOP is associated with multiple runtime variants that instantiate different runtime inputs and execution conditions while keeping the task specification and reference workflow fixed. A runtime variant contains a public runtime input together with the environment configuration required to execute that input. The environment configuration may include hidden connector state and controlled tool failures.

Most benchmark variants were generated through the model-based variant-construction pipeline. Nine additional variants were constructed through coverage synthesis. The accepted variants are stored as part of the released benchmark.

\textbf{Control-flow coverage.}
Variant construction tracks coverage obligations derived from the reference workflow. For each conditional, the construction targets both branch outcomes. For each try/catch block, it targets execution both without an exception and through the catch path. For loops, it targets zero, one, and multiple iterations when these behaviors are represented by the workflow and variant construction.

\textbf{Fault injection.}
Variants may include a controlled failure for an operation inside an error-handling region. A fault is matched to an executed operation using the tool name together with a literal argument selector when available. If this does not uniquely identify the target call, matching falls back to the ordinal occurrence of that tool. The same fault registry and matching procedure are used when executing different orchestration protocols.

\textbf{Execution isolation.}
Before each execution, the corresponding connector state is initialized from the runtime variant. Each execution uses an independent state, and the environment is reset before the next execution.

\subsection{Evaluation Details}
\label{app:evaluation_details}

All orchestration protocols are scored from their realized executions using the same evaluator. For each runtime variant, the evaluator receives the executed tool trajectory, execution failures, resulting simulated connector state, externally recorded effects where applicable, and terminal output.

Reference workflows represent the prescribed enterprise SOPs rather than arbitrary implementations of underspecified tasks. Accordingly, evaluation measures fidelity to the prescribed workflow behavior, including its required operation order, as well as the resulting state and outputs.

\textbf{Why evaluate fidelity to the prescribed SOP.} Enterprise SOPs prescribe the procedure as well as the outcome. For example, an approval may need to precede a payment, or a duplicate check may need to precede record creation. An execution that reaches the correct final state through a different procedure can still violate compliance or audit requirements, so we treat the prescribed procedure as part of the task. The required order is not hidden from the model: task descriptions retain the ordering constraints of the source SOP (Appendix~\ref{app:bench_task_desc}). Execution Trace Accuracy contributes only half of Orchestration Accuracy; the other half, Side-Effect Accuracy, evaluates the resulting state. Final Accuracy evaluates only side effects and terminal output, independent of the execution trajectory. The evaluator further normalizes runtime-generated identifiers and compares supported queries by result semantics rather than exact form.

\textbf{Execution Trace Accuracy.}
Execution Trace Accuracy is the mean of five binary components: tool selection, tool order, tool execution, required arguments, and control path.

Tool selection requires the executed tool identities and call counts to match the reference SOP. Tool order measures whether the realized execution follows the prescribed operation order of the reference SOP and requires equality between the executed tool-name sequence and the reference tool-name sequence. Tool execution requires tool selection and tool order to pass and additionally requires that no unexpected execution failure occur. Required arguments evaluates the required arguments associated with the reference tool calls. Control path checks whether the realized execution follows the prescribed branch, error-handling, and termination behavior.

These components are evaluated jointly from the realized execution. For example, an additional tool call may affect tool selection, tool order, tool execution, and control-path agreement.

\textbf{Required arguments.}
Argument scoring is performed on the arguments of executed tool calls after runtime values have been resolved. Runtime-generated identifiers covered by the benchmark normalization procedure are normalized before comparison. For standard tool arguments, the evaluator checks the required argument fields against the corresponding reference values.

The evaluator includes operation-specific handling for supported Salesforce and Snowflake query arguments, where query behavior can be compared using result or state semantics.

\textbf{Side-Effect Accuracy.}
Side-Effect Accuracy is binary and is evaluated over the state produced in the simulated connector backend together with externally recorded effects used by the corresponding verifier. The verifier checks that required changes are present and that prohibited or unintended changes covered by the verifier are absent.

Side-Effect Accuracy evaluates the resulting state rather than the complete sequence of intermediate connector states. This complements Execution Trace Accuracy, which evaluates the realized tool trajectory.

\textbf{Termination and partial executions.}
Executions that terminate because of a tool failure, formatting limit, or execution budget are scored using the trajectory and simulated connector state accumulated before termination.

\textbf{Aggregation.}
Each of the five Execution Trace Accuracy components is binary, and their mean forms the per-variant Execution Trace Accuracy. Side-Effect Accuracy is binary. Orchestration Accuracy is the average of Execution Trace Accuracy and Side-Effect Accuracy for the same runtime variant. We then average the per-variant score across the benchmark. Terminal-output accuracy is reported separately.

\subsection{Training Data Construction}
\label{app:training_data}

We construct the training data from a separate set of 3,000 source recipes. Recipes are separated from the \textsc{AgentOrch} benchmark by source recipe ID.

For each source recipe, we construct a task specification following the same task-description procedure used for the benchmark. The source recipe serves as the reference workflow. We deterministically convert the complete reference workflow into the serialized workflow target used for training. Each recipe produces one training example; concrete runtime variants are not included in the training input and do not specialize the target.

\textbf{Complete workflow targets.}
The training target retains the full workflow structure represented by the source recipe. This includes both outcomes of conditional branches, loop nodes and their loop bodies, try and catch regions, tool calls, dependencies on runtime inputs and preceding tool results, and terminal control behavior. Loops are represented as workflow structure rather than being expanded into the number of iterations observed in one runtime execution.

\textbf{Target validation.}
Each target is parsed and normalized before inclusion. The validation checks the workflow syntax and structure, references between workflow nodes, availability of the tools used by the target, and the argument schemas of those tools. Compilation, schema, planner-protocol, and tool-surface checks are fail-closed.

These checks do not execute the serialized training target on runtime variants. Replay of the original source recipe is not an inclusion criterion for the active 3,000-example training set. We therefore use structural and interface validation, rather than execution validation, to describe the validation of the training targets.

\textbf{Tool exposure.}
Across the training corpus, the tool-card pool contains 1,036 distinct tool cards spanning 306 namespaces. Of these, 1,033 tool cards spanning 305 namespaces occur in the reference workflow targets. Each individual training example exposes 64 tool cards. The 64-card set contains the tools required by that recipe together with sampled distractor tools and therefore varies across training examples.

\textbf{Train--benchmark separation.}
Training and benchmark sets are disjoint at the source-recipe level: no source recipe used in the AGENTORCH benchmark is used for training. Both sets are drawn from the same enterprise automation environment and therefore intentionally share the same tool ecosystem and may contain recurring workflow motifs. The split is designed to evaluate generalization to unseen source recipes and task instances, rather than transfer to unseen tools or a different automation domain. Thus, all reported training results are evaluated on source recipes that are not observed during training.

\subsection{Training Details}
\label{app:training_details}

\textbf{Supervised Fine-Tuning.}
We apply supervised fine-tuning to five base models using the training pairs described in Appendix~\ref{app:training_data}. The model receives the task specification and is trained to generate the complete reference workflow. The loss is applied only to the workflow response; tokens in the input prompt are masked from the training loss. We do not supervise chain-of-thought explanations or intermediate reasoning. All reported SFT runs use LoRA adaptation.

\textbf{Reinforcement Learning.}
We further train models using Group Relative Policy Optimization (GRPO). The reward is computed by comparing a sampled workflow directly with the serialized reference workflow. Reward computation does not execute the sampled workflow and does not invoke either connector backend.

A sampled output must first be parsed as a workflow. If parsing fails, the total reward is zero. For valid outputs, the reward combines five components:

\begin{itemize}[leftmargin=*]
    \item \textbf{Validity} (0.15): whether the sampled response can be parsed as a valid workflow.
    
    \item \textbf{Tool calls} (0.25): we recursively flatten tool-call nodes from the generated and reference workflows and compute the longest-common-subsequence similarity between their tool-name sequences.
    
    \item \textbf{Arguments} (0.35): tool calls are aligned using the tool-name sequence alignment, after which their arguments are compared after canonical JSON serialization. The component is normalized by the number of reference tool calls.
    
    \item \textbf{Workflow structure} (0.15): we flatten the workflow-node types, including call, conditional, loop, try/catch, transformation, and termination nodes, and compute longest-common-subsequence similarity between the generated and reference node-type sequences.
    
    \item \textbf{Exact match} (0.10): the complete generated workflow receives this component when its canonical JSON serialization exactly matches the reference target.
\end{itemize}

The reward contains no execution-based component. In particular, it does not use connector observations, simulated state, Side-Effect Accuracy, or the benchmark evaluator. It also contains no separate response-length reward. Outputs that cannot be parsed, including outputs truncated before a valid workflow is completed, receive zero reward through the validity check.

For GRPO, we sample eight responses per prompt and normalize advantages within each group. The RL dataset uses 1,000 training prompts selected by prompt length to fit the configured generation limits. Model-specific LoRA ranks, learning rates, KL coefficients, generation limits, and optimization settings are provided in Table~\ref{tab:training_hparams}.

\begin{table}[t]
\centering
\small
\setlength{\tabcolsep}{3.5pt}
\caption{\small Training hyperparameters.}
\label{tab:training_hparams}
\begin{tabular}{lccc}
\toprule
\textbf{Setting} & \textbf{SFT} & \textbf{Qwen RL} & \textbf{Gemma RL} \\
\midrule
Method
& LoRA ($r$=16, $\alpha$=32)
& GRPO
& GRPO + LoRA ($r$=16, $\alpha$=32) \\

Learning rate
& $1\times10^{-4}$
& $3\times10^{-6}$ (27B)
& $1\times10^{-5}$ \\

Batch / rollouts
& 16 seq.
& 16 / 8 rollouts
& 8 prompts $\times$ 8 gen. \\

Max. length
& 81,920
& 65,536 / 6,144
& 4,096 completion \\

KL coefficient
& -- & 0.001 & 0.001 \\
\bottomrule
\end{tabular}
\end{table}

\section{More Experimental Analysis}
\subsection{Full Results for RQ1}
\label{app:rq1_full}

We provide full results of our RQ1 below:
\begin{table}[t]
    \centering
    \caption{
    \textbf{Scaling does not consistently narrow the
    \textsc{Global}--\textsc{Step-by-Step} gap.}
    We report performance under the \textsc{Global} and
    \textsc{Step-by-Step} (\texttt{react}) settings.
    $\Delta$ denotes \textsc{Step-by-Step} minus \textsc{Global}
    in percentage points (pp).
    }
    \label{tab:rq1_scaling}

    \scriptsize
    \setlength{\tabcolsep}{3.2pt}
    \renewcommand{\arraystretch}{1.08}

    \resizebox{\linewidth}{!}{
    \begin{tabular}{
        @{} l c
        c c c
        c c c
        c c c @{}
    }
    \toprule

    & &
    \multicolumn{3}{c}{Orchestration Accuracy} &
    \multicolumn{3}{c}{Final Accuracy} &
    \multicolumn{3}{c}{Error Rate} \\

    \cmidrule(lr){3-5}
    \cmidrule(lr){6-8}
    \cmidrule(lr){9-11}

    Family & Scale &
    Global & Step-by-Step & $\Delta$ &
    Global & Step-by-Step & $\Delta$ &
    Global & Step-by-Step & $\Delta$ \\

    \midrule

    Qwen2.5
    & 0.5B
    & 0.13\% & 36.46\% & +36.33
    & 0.00\% & 12.45\% & +12.45
    & 99.78\% & 0.22\% & $-99.56$ \\

    & 1.5B
    & 1.66\% & 31.41\% & +29.75
    & 0.22\% & 8.91\% & +8.69
    & 96.43\% & 0.25\% & $-96.18$ \\

    & 7B
    & 2.63\% & 30.50\% & +27.87
    & 0.47\% & 7.42\% & +6.95
    & 93.45\% & 1.09\% & $-92.36$ \\

    & 14B
    & 5.33\% & 41.54\% & +36.21
    & 1.61\% & 21.46\% & +19.85
    & 85.40\% & 0.03\% & $-85.37$ \\

    & 32B
    & 16.18\% & 45.64\% & +29.46
    & 6.40\% & 24.41\% & +18.01
    & 58.14\% & 0.34\% & $-57.80$ \\

    & 72B
    & 9.91\% & 46.57\% & +36.66
    & 3.91\% & 25.96\% & +22.05
    & 73.57\% & 0.25\% & $-73.32$ \\

    \midrule

    Gemma-3
    & 4B
    & 6.15\% & 32.75\% & +26.60
    & 1.61\% & 10.65\% & +9.04
    & 82.39\% & 0.28\% & $-82.11$ \\

    & 12B
    & 8.46\% & 37.85\% & +29.39
    & 1.83\% & 13.07\% & +11.24
    & 77.17\% & 0.03\% & $-77.14$ \\

    & 27B
    & 7.41\% & 39.93\% & +32.52
    & 2.67\% & 17.02\% & +14.35
    & 78.14\% & 0.09\% & $-78.05$ \\

    \bottomrule
    \end{tabular}
    }
\end{table}

\begin{table*}[t]
\vspace{-2mm}
\centering
\caption{\small
\textbf{Reasoning does not consistently improve orchestration performance.}
We compare base and thinking variants under both
\textsc{Global} and \textsc{Step-by-Step} (\texttt{react}) execution.
Within each base--thinking pair and execution setting, the better-performing
score is highlighted and bolded (higher is better for accuracy metrics;
lower is better for error rate).
Mean accuracy is the average of orchestration, final, side-effect,
execution-trace, and terminal-output accuracy.
}
\label{tab:rq1_reasoning_add}

\scriptsize
\setlength{\tabcolsep}{2.2pt}
\renewcommand{\arraystretch}{1.05}

\textbf{(a) \textsc{Global}}
\vspace{1mm}

\resizebox{\textwidth}{!}{%
\begin{tabular}{@{}l
                rr@{\hspace{2pt}\vrule\hspace{2pt}}
                rr@{\hspace{2pt}\vrule\hspace{2pt}}
                rr@{\hspace{2pt}\vrule\hspace{2pt}}
                rr@{\hspace{2pt}\vrule\hspace{2pt}}
                rr@{\hspace{2pt}\vrule\hspace{2pt}}
                rr@{\hspace{2pt}\vrule\hspace{2pt}}
                rr@{\hspace{2pt}\vrule\hspace{2pt}}
                rr@{}}

\toprule

& \multicolumn{2}{c}{\textbf{Qwen3-4B}}
& \multicolumn{2}{c}{\textbf{Qwen3-30B}}
& \multicolumn{2}{c}{\textbf{Qwen3.8-27B}}
& \multicolumn{2}{c}{\textbf{DeepSeek-V4}}
& \multicolumn{2}{c}{\textbf{GLM-4.5-Air}}
& \multicolumn{2}{c}{\textbf{GLM-4.5-FP8}}
& \multicolumn{2}{c}{\textbf{GLM-5.2-FP8}}
& \multicolumn{2}{c}{\textbf{Claude-Opus-5}}
\\

\cmidrule(lr){2-3}
\cmidrule(lr){4-5}
\cmidrule(lr){6-7}
\cmidrule(lr){8-9}
\cmidrule(lr){10-11}
\cmidrule(lr){12-13}
\cmidrule(lr){14-15}
\cmidrule(lr){16-17}

\textbf{Metric}
& Base & Think
& Base & Think
& Base & Think
& Base & Think
& Base & Think
& Base & Think
& Base & Think
& Base & Think
\\

\midrule

Orch.
& \best{13.75\%} & 10.25\%
& 8.35\% & \best{14.75\%}
& 28.57\% & \best{33.02\%}
& \best{27.01\%} & 17.14\%
& \best{12.28\%} & 10.72\%
& 12.93\% & \best{20.09\%}
& \best{19.23\%} & 17.75\%
& 46.44\% & \best{59.88\%}
\\

Final
& \best{3.91\%} & 2.70\%
& 2.95\% & \best{5.19\%}
& 12.67\% & \best{17.05\%}
& \best{11.43\%} & 9.66\%
& \best{4.78\%} & 4.25\%
& 5.19\% & \best{8.48\%}
& 8.57\% & \best{9.29\%}
& 23.54\% & \best{32.33\%}
\\

Side
& \best{16.40\%} & 13.04\%
& 9.44\% & \best{16.83\%}
& 31.80\% & \best{35.71\%}
& \best{29.94\%} & 17.52\%
& \best{14.22\%} & 11.61\%
& 13.88\% & \best{21.46\%}
& \best{21.15\%} & 18.66\%
& 50.16\% & \best{63.98\%}
\\

Trace
& \best{11.11\%} & 7.45\%
& 7.26\% & \best{12.68\%}
& 25.33\% & \best{30.33\%}
& \best{24.07\%} & 16.76\%
& \best{10.34\%} & 9.83\%
& 11.99\% & \best{18.72\%}
& \best{17.31\%} & 16.84\%
& 42.72\% & \best{55.78\%}
\\

Term.
& \best{16.30\%} & 8.42\%
& 13.73\% & \best{20.12\%}
& 28.98\% & \best{35.16\%}
& \best{28.98\%} & 20.06\%
& 11.71\% & \best{13.45\%}
& 10.81\% & \best{19.88\%}
& \best{18.94\%} & 14.32\%
& 42.24\% & \best{48.29\%}
\\

\midrule

\textbf{Mean Acc. $\uparrow$}
& \best{12.29\%} & 8.37\%
& 8.35\% & \best{13.91\%}
& 25.47\% & \best{30.25\%}
& \best{24.29\%} & 16.23\%
& \best{10.67\%} & 9.97\%
& 10.96\% & \best{17.73\%}
& \best{17.04\%} & 15.37\%
& 41.02\% & \best{52.05\%}
\\

\textbf{Error $\downarrow$}
& \best{61.96\%} & 77.76\%
& 76.43\% & \best{60.81\%}
& 43.88\% & \best{39.47\%}
& \best{43.51\%} & 67.67\%
& \best{73.76\%} & 74.63\%
& 75.68\% & \best{58.35\%}
& \best{61.34\%} & 73.01\%
& 19.16\% & \best{6.21\%}
\\

\bottomrule
\end{tabular}%
}

\vspace{2.5mm}

\textbf{(b) \textsc{Step-by-Step} (\texttt{react})}
\vspace{1mm}

\resizebox{\textwidth}{!}{%
\begin{tabular}{@{}l
                rr@{\hspace{2pt}\vrule\hspace{2pt}}
                rr@{\hspace{2pt}\vrule\hspace{2pt}}
                rr@{\hspace{2pt}\vrule\hspace{2pt}}
                rr@{\hspace{2pt}\vrule\hspace{2pt}}
                rr@{\hspace{2pt}\vrule\hspace{2pt}}
                rr@{\hspace{2pt}\vrule\hspace{2pt}}
                rr@{\hspace{2pt}\vrule\hspace{2pt}}
                rr@{}}

\toprule

& \multicolumn{2}{c}{\textbf{Qwen3-4B}}
& \multicolumn{2}{c}{\textbf{Qwen3-30B}}
& \multicolumn{2}{c}{\textbf{Qwen3.8-27B}}
& \multicolumn{2}{c}{\textbf{DeepSeek-V4}}
& \multicolumn{2}{c}{\textbf{GLM-4.5-Air}}
& \multicolumn{2}{c}{\textbf{GLM-4.5-FP8}}
& \multicolumn{2}{c}{\textbf{GLM-5.2-FP8}}
& \multicolumn{2}{c}{\textbf{Claude-Opus-5}}
\\

\cmidrule(lr){2-3}
\cmidrule(lr){4-5}
\cmidrule(lr){6-7}
\cmidrule(lr){8-9}
\cmidrule(lr){10-11}
\cmidrule(lr){12-13}
\cmidrule(lr){14-15}
\cmidrule(lr){16-17}

\textbf{Metric}
& Base & Think
& Base & Think
& Base & Think
& Base & Think
& Base & Think
& Base & Think
& Base & Think
& Base & Think
\\

\midrule

Orch.
& 41.84\% & \best{47.34\%}
& 47.81\% & \best{52.09\%}
& 53.31\% & \best{63.08\%}
& \best{58.99\%} & 58.85\%
& 48.59\% & \best{49.49\%}
& 54.47\% & \best{54.68\%}
& 62.16\% & \best{62.66\%}
& 63.52\% & \best{68.41\%}
\\

Final
& 17.30\% & \best{22.17\%}
& 26.68\% & \best{28.17\%}
& 31.77\% & \best{49.41\%}
& 44.57\% & \best{45.43\%}
& 28.88\% & \best{30.06\%}
& \best{38.29\%} & 37.92\%
& 48.20\% & \best{48.91\%}
& 47.14\% & \best{55.31\%}
\\

Side
& 51.37\% & \best{57.05\%}
& 58.57\% & \best{63.35\%}
& 64.91\% & \best{72.39\%}
& \best{68.26\%} & 67.67\%
& 59.69\% & \best{59.88\%}
& \best{64.13\%} & 64.10\%
& 71.58\% & \best{72.08\%}
& 70.09\% & \best{74.57\%}
\\

Trace
& 32.32\% & \best{37.62\%}
& 37.04\% & \best{40.82\%}
& 41.71\% & \best{53.76\%}
& 49.73\% & \best{50.03\%}
& 37.48\% & \best{39.11\%}
& 44.81\% & \best{45.26\%}
& 52.73\% & \best{53.24\%}
& 56.95\% & \best{62.25\%}
\\

Term.
& 52.17\% & \best{53.70\%}
& \best{57.24\%} & 56.55\%
& 60.06\% & \best{72.42\%}
& 71.21\% & \best{71.43\%}
& \best{59.88\%} & 58.14\%
& \best{67.89\%} & 67.08\%
& \best{72.98\%} & 72.48\%
& 72.36\% & \best{77.55\%}
\\

\midrule

\textbf{Mean Acc. $\uparrow$}
& 39.00\% & \best{43.58\%}
& 45.47\% & \best{48.20\%}
& 50.35\% & \best{62.21\%}
& 58.55\% & \best{58.68\%}
& 46.90\% & \best{47.34\%}
& \best{53.92\%} & 53.81\%
& 61.53\% & \best{61.87\%}
& 62.01\% & \best{67.62\%}
\\

\textbf{Error $\downarrow$}
& \best{0.06\%} & 0.16\%
& 0.19\% & \best{0.06\%}
& \best{0.16\%} & 0.71\%
& \best{0.19\%} & 1.68\%
& \best{0.12\%} & 4.04\%
& \best{0.09\%} & 0.19\%
& \best{0.12\%} & 0.53\%
& \best{0.00\%} & \best{0.00\%}
\\

\bottomrule
\end{tabular}%
}

\vspace{-4mm}
\end{table*}

\subsection{Additional Analyses for RQ3}
\label{app:rq3_additional}

We examine how feedback scope affects \textsc{Replan} gains, how these gains extend to held-out runtime variants, and which failures are resolved by workflow revision. Mean Accuracy averages Orchestration, Final, Side-Effect, Execution Trace, and Terminal Output Accuracy.

\textbf{Two-Variant Feedback Produces Similar Average Gains to All-Variant Feedback.}
We compare feedback from two variants of each SOP with feedback from all variants, evaluating both settings on all variants. Two-variant feedback improves Mean Accuracy by 5.49 pp on average, 0.05 pp more than all-variant feedback. It reduces Error Rate by 12.07 pp, 0.17 pp less than all-variant feedback (Table~\ref{tab:rq2_feedback_scope}).

\begin{table}[t]
\centering
\caption{\small
\textbf{Effect of limiting \textsc{Replan} feedback to two runtime variants per SOP.}
All columns evaluate on all runtime variants. Gaps compare two-variant feedback with all-variant feedback. Error reduction is $\text{Error}_{R0}-\text{Error}_{R3}$. All values are in pp.
}
\label{tab:rq2_feedback_scope}
\small
\setlength{\tabcolsep}{5pt}
\renewcommand{\arraystretch}{1.06}
\resizebox{\linewidth}{!}{
\begin{tabular}{lrrrr}
\toprule
\textbf{Model}
& \textbf{Two-Var. $\Delta$Mean Acc.}
& \textbf{Gap vs. All-Var.}
& \textbf{Two-Var. Error Red.}
& \textbf{Gap vs. All-Var.} \\
\midrule
Qwen3.8-27B             & +10.42 & +0.21 & 14.00 & -0.53 \\
DeepSeek-V4-Flash       & +6.88  & -0.02 & 13.32 & -0.07 \\
Mistral-Small-3.1-24B   & +7.02  & +0.16 & 17.08 & +0.28 \\
Gemma-4-26B-A4B         & +5.12  & +0.27 & 11.55 & -0.62 \\
Gemma-4-12B             & +4.70  & +0.50 & 12.52 & +1.31 \\
Qwen3-4B                & +4.13  & -0.02 & 13.14 & -0.34 \\
Qwen3-30B-A3B           & +2.72  & -0.73 & 5.81  & -1.37 \\
Gemma-3-27B             & +2.90  & +0.03 & 9.16  & +0.00 \\
\midrule
\textbf{Average}        & \textbf{+5.49} & \textbf{+0.05}
                        & \textbf{12.07}  & \textbf{-0.17} \\
\bottomrule
\end{tabular}
}
\end{table}

\textbf{Held-Out Variants Improve as Much as Feedback Variants.}
Under two-variant feedback, we compare the improvement on the variants that provided feedback with that on held-out variants. The average held-out minus seen difference is $+0.03$ pp for Mean Accuracy and $-0.31$ pp for Error Rate reduction (Table~\ref{tab:rq2_seen_gap}). The revised workflow thus improves as much on variants it never observed as on those used for feedback.

\begin{table}[t]
\centering
\caption{\small
\textbf{Transfer from feedback variants to held-out variants.}
\textsc{Replan} feedback uses two variants of each SOP. The transfer gap is the held-out value minus the seen value; values near zero indicate similar improvement on both subsets.
}
\label{tab:rq2_seen_gap}
\small
\setlength{\tabcolsep}{5pt}
\renewcommand{\arraystretch}{1.06}
\resizebox{\linewidth}{!}{
\begin{tabular}{lrrrr}
\toprule
\textbf{Model}
& \textbf{Seen $\Delta$Mean Acc.}
& \textbf{Held-out--Seen Gap}
& \textbf{Seen Error Red.}
& \textbf{Held-out--Seen Gap} \\
\midrule
Qwen3.8-27B             & +10.81 & -0.65 & 14.55 & -0.91 \\
DeepSeek-V4-Flash       & +7.07  & -0.31 & 12.92 & +0.67 \\
Mistral-Small-3.1-24B   & +7.59  & -0.96 & 18.13 & -1.74 \\
Gemma-4-26B-A4B         & +5.38  & -0.43 & 12.29 & -1.23 \\
Gemma-4-12B             & +4.45  & +0.42 & 12.92 & -0.67 \\
Qwen3-4B                & +3.63  & +0.83 & 13.16 & -0.03 \\
Qwen3-30B-A3B           & +2.32  & +0.66 & 5.30  & +0.85 \\
Gemma-3-27B             & +2.52  & +0.64 & 8.79  & +0.61 \\
\midrule
\textbf{Average}        & \textbf{+5.47} & \textbf{+0.03}
                        & \textbf{12.26} & \textbf{-0.31} \\
\bottomrule
\end{tabular}
}
\end{table}

\textbf{Gains Also Hold on Two Held-Out Variants per SOP.}
Evaluating on a subset of two held-out variants per SOP, Mean Accuracy improves for all eight models by 5.34 pp on average, and Error Rate decreases by 12.31 pp (Table~\ref{tab:rq2_heldout2}). The transfer result therefore does not depend on aggregating over all held-out variants.

\begin{table}[t]
\centering
\caption{\small
\textbf{\textsc{Replan} performance on a subset of two held-out variants per SOP.}
Feedback uses two variants of each SOP; evaluation uses two other variants never used for feedback. Base is Round 0 and Final is Round 3.
}
\label{tab:rq2_heldout2}
\small
\setlength{\tabcolsep}{4.5pt}
\renewcommand{\arraystretch}{1.06}
\resizebox{0.68\linewidth}{!}{
\begin{tabular}{lrrrrrr}
\toprule
\textbf{Model}
& \multicolumn{3}{c}{\textbf{Mean Accuracy} $\uparrow$}
& \multicolumn{3}{c}{\textbf{Error Rate} $\downarrow$} \\
\cmidrule(lr){2-4}
\cmidrule(lr){5-7}
& \textbf{Base} & \textbf{Final} & $\boldsymbol{\Delta}$
& \textbf{Base} & \textbf{Final} & $\boldsymbol{\Delta}$ \\
\midrule
Qwen3.8-27B             & 26.27 & 36.13 & +9.86 & 42.96 & 28.43 & -14.53 \\
DeepSeek-V4-Flash       & 24.72 & 31.05 & +6.32 & 42.87 & 29.69 & -13.18 \\
Mistral-Small-3.1-24B   & 13.31 & 20.40 & +7.09 & 67.35 & 48.79 & -18.56 \\
Gemma-4-26B-A4B         & 4.12  & 9.20  & +5.09 & 88.88 & 77.31 & -11.57 \\
Gemma-4-12B             & 12.17 & 16.89 & +4.72 & 65.65 & 53.09 & -12.56 \\
Qwen3-4B                & 12.80 & 16.92 & +4.12 & 60.81 & 47.26 & -13.55 \\
Qwen3-30B-A3B           & 8.32  & 10.93 & +2.61 & 75.96 & 70.67 & -5.29 \\
Gemma-3-27B             & 6.93  & 9.83  & +2.91 & 79.10 & 69.87 & -9.23 \\
\midrule
\textbf{Average}        & \textbf{13.58} & \textbf{18.92} & \textbf{+5.34}
                        & \textbf{65.45} & \textbf{53.14} & \textbf{-12.31} \\
\bottomrule
\end{tabular}
}
\end{table}

\textbf{Improvements Are Widespread Across Held-Out Variants.}
To check whether the average gain is driven by a few variants, we count held-out variants whose score changes from Round 0 to Round 3. For every model, improvements outnumber regressions on both Orchestration and Final Accuracy. In total across models, Orchestration Accuracy improves on 2,214 held-out variants and decreases on 165, while Final Accuracy improves on 555 and decreases on 75 (Table~\ref{tab:rq2_row_changes}).

\begin{table}[t]
\centering
\caption{\small
\textbf{Variant-level improvements and regressions on held-out variants.}
Counts of held-out variants whose score increases (Up) or decreases (Down) from Round 0 to Round 3 under two-variant feedback. Unchanged variants are omitted.
}
\label{tab:rq2_row_changes}
\small
\setlength{\tabcolsep}{5.5pt}
\renewcommand{\arraystretch}{1.06}
\begin{tabular}{lrrrr}
\toprule
\textbf{Model}
& \textbf{Orch. Up}
& \textbf{Orch. Down}
& \textbf{Final Up}
& \textbf{Final Down} \\
\midrule
Qwen3.8-27B             & 428 & 31 & 161 & 9 \\
DeepSeek-V4-Flash       & 335 & 51 & 110 & 26 \\
Mistral-Small-3.1-24B   & 357 & 31 & 68  & 12 \\
Gemma-4-26B-A4B         & 240 & 12 & 56  & 10 \\
Gemma-4-12B             & 264 & 6  & 49  & 4 \\
Qwen3-4B                & 249 & 12 & 41  & 6 \\
Qwen3-30B-A3B           & 155 & 12 & 43  & 5 \\
Gemma-3-27B             & 186 & 10 & 27  & 3 \\
\midrule
\textbf{Total}          & \textbf{2214} & \textbf{165}
                        & \textbf{555} & \textbf{75} \\
\bottomrule
\end{tabular}
\end{table}

\textbf{Most Repaired Workflows Initially Fail Schema, Type, or Reference Checks.}
Across the eight models, 641 workflows are invalid at Round 0 and become valid by Round 3. Of these, 538 (83.9\%) initially parse successfully but fail schema, type, or reference validation. These checks cover constraints on tool arguments and references between workflow steps. The remaining 103 workflows (16.1\%) fail response-parsing or plan-structure checks (Table~\ref{tab:replan_static_detection}). The share of parsing or plan-structure failures is highest for Gemma-4-26B-A4B, at 39.8\%. Together, these checks detect the initial failures in all 641 repaired workflows before execution.

\begin{table}[t]
\centering
\caption{\small
\textbf{Failure types of invalid workflows repaired by \textsc{Replan}.}
Repaired workflows are invalid at Round 0 and valid by Round 3. Parse/Shape: failures in response parsing or plan-structure checks. Validator: workflows that parse successfully but fail schema, type, or reference validation. Together, these checks detect all 641 repaired workflows before execution.
}
\label{tab:replan_static_detection}
\small
\setlength{\tabcolsep}{4pt}
\renewcommand{\arraystretch}{1.06}
\begin{tabular}{lrrrr}
\toprule
\textbf{Model}
& \textbf{Repaired}
& \textbf{Parse / Shape}
& \textbf{Validator}
& \textbf{Validator (\%)} \\
\midrule
Qwen3.8-27B              & 97  & 16 & 81  & 83.5 \\
DeepSeek-V4-Flash        & 84  & 12 & 72  & 85.7 \\
Mistral-Small-3.1-24B    & 117 & 13 & 104 & 88.9 \\
Gemma-4-12B              & 79  & 9  & 70  & 88.6 \\
Qwen3-4B                 & 85  & 13 & 72  & 84.7 \\
Qwen3-30B-A3B            & 40  & 5  & 35  & 87.5 \\
Gemma-3-27B              & 56  & 2  & 54  & 96.4 \\
Gemma-4-26B-A4B          & 83  & 33 & 50  & 60.2 \\
\midrule
\textbf{All models}      & \textbf{641} & \textbf{103}
                         & \textbf{538} & \textbf{83.9} \\
\bottomrule
\end{tabular}
\end{table}

\paragraph{Summary.}
\textsc{Replan} gains require feedback from only two variants per SOP, extend equally to variants never used for feedback, and are spread across most held-out variants rather than concentrated in a few. The invalid workflows it repairs mostly parse correctly but specify tool arguments or references between steps incorrectly, and all of these defects are detectable before execution. Consistent with the main text, these results identify workflow validity, particularly correct arguments and references, as a major bottleneck in global orchestration.

\section{Additional Related Work}
\label{app:rel_wok}
\textbf{Workflow generation and process automation.}
The closest work studies workflow generation and automation. WorfBench~\citep{qiao2025benchmarking}, WorkflowLLM~\citep{fan2025workflowllm}, AutoFlow~\citep{li2024autoflow}, AFlow~\citep{zhang2025aflow}, FlowMind~\citep{zeng2023flowmind}, Chat2Workflow~\citep{zhong2026chat2workflow}, and AutomationBench~\citep{shepard2026automationbench} consider agentic pipelines, visual workflows, RPA-style procedures, or cross-application automation. Earlier work studies IFTTT-style trigger-action programming~\citep{quirk2015language,chaurasia2017dialog,ur2016trigger}, while recent business-process research explores language-based process modeling and generation~\citep{kourani2024process,busch2024towards,kourani2025evaluating,li2025llm}. Our work differs by treating complete workflow generation as the \textsc{Global} orchestration paradigm and directly comparing it with \textsc{Step-by-Step} interaction on the same executable enterprise workflows, including connector contracts, parameter bindings, dependencies, transformations, branching, looping, and observable side effects. We further study how global orchestration can be improved through execution feedback and SOP-grounded supervision.

\section{Simulator Fidelity via Reference-Action Replay}
\label{app:simulator_fidelity}

Model-generated workflows in \textsc{AgentOrch} are evaluated using the
simulated connector backend, while reference executions are collected from
the corresponding real connectors where supported. We therefore validate
whether the simulator preserves the benchmark-relevant behavior observed in
the real-backend reference executions.

Specifically, we replay the recorded reference tool calls from each
real-backend execution one action at a time in the simulated backend, without
any model planning. The resulting executions are scored using the same
evaluation pipeline as model-generated runs, including the same normalization
of dynamic identifiers, timestamps, argument values, and query semantics.

Table~\ref{tab:simulator_fidelity} reports the agreement between these
simulated replays and the stored real-backend reference executions. The
simulator achieves 97.20\% agreement on tool selection, tool order, tool
execution, and control path, 99.97\% on required arguments, 97.75\%
Execution Trace Accuracy, and 99.32\% Side-Effect Accuracy. The resulting
Orchestration Accuracy is 98.54\%.

\begin{table}[t]
    \centering
    \caption{\small
        Simulator fidelity under reference-action replay.
        Recorded tool calls from the real-backend reference executions are
        replayed one action at a time in the simulated backend and evaluated
        using the same benchmark evaluator as model-generated runs.
    }
    \label{tab:simulator_fidelity}
    \small
    \begin{tabular}{lc}
        \toprule
        \textbf{Metric} & \textbf{Reference-Action Replay} \\
        \midrule
        Tool Selection              & 97.20\% \\
        Tool Order                  & 97.20\% \\
        Tool Execution              & 97.20\% \\
        Required Arguments          & 99.97\% \\
        Control Path                & 97.20\% \\
        Execution Trace Accuracy    & 97.75\% \\
        Side-Effect Accuracy        & 99.32\% \\
        Orchestration Accuracy      & 98.54\% \\
        \bottomrule
    \end{tabular}
    
\end{table}

These results show that the simulated backend reproduces the
benchmark-relevant execution behavior of the recorded real-backend references
with high fidelity. This makes broad simulator--reference mismatch an
unlikely explanation for the observed performance gap between
\textsc{Global} and \textsc{Step-by-Step} orchestration.

\section{Performance Across Workflow Characteristics}
\label{app:workflow_characteristics}

We next examine whether the gap between \textsc{Global} and
\textsc{Step-by-Step} is concentrated in particular types of workflows.
We characterize each SOP using its reference workflow. Workflow size is
measured by the number of tool-call nodes, with short, medium, and long
workflows defined as 1--2, 3--5, and at least 6 calls, respectively, based
on the quartiles of the benchmark distribution. We additionally record
whether the workflow contains conditional branches, loops, or try/catch
regions, and measure its maximum nesting depth. To characterize
cross-application workflows, we count the number of distinct application
families used by tool-call nodes in the reference workflow, excluding
runtime-helper tools. Fault injection is defined at the runtime-variant
level.

We report results over the 31 models for which both \textsc{Global} and
\textsc{Step-by-Step} evaluations are complete. Fine-tuned checkpoints are
excluded from this analysis. Failed executions remain in the evaluation and
receive zero accuracy, following the main benchmark evaluation. The gap is
defined as \textsc{Step-by-Step} Orchestration Accuracy minus
\textsc{Global} Orchestration Accuracy. Confidence intervals are computed
using a bootstrap clustered by SOP.

\begin{table*}[t]
\centering
\caption{\small
Performance across workflow characteristics.
Orchestration Accuracy and Error Rate are reported in percent.
The gap is \textsc{Step-by-Step} minus \textsc{Global} Orchestration
Accuracy in percentage points. Confidence intervals are 95\% bootstrap
intervals clustered by SOP. Workflow characteristics are derived from the
reference workflow, except fault injection, which is defined per runtime
variant.
}
\label{tab:workflow_characteristics}
\small
\resizebox{\textwidth}{!}{
\begin{tabular}{lrrrrrrrr}
\toprule
\textbf{Subset}
& \textbf{\# SOP}
& \textbf{\# Var.}
& \textbf{GLOBAL}
& \textbf{STEP-BY-STEP}
& \textbf{Gap}
& \textbf{95\% CI}
& \textbf{GLOBAL Err.}
& \textbf{STEP Err.} \\
\midrule

\multicolumn{9}{l}{\textit{All workflows}} \\
All
& 650 & 3220 & 13.8 & 46.4 & +32.6
& [31.2, 34.2] & 70.7 & 0.5 \\

\midrule
\multicolumn{9}{l}{\textit{Workflow size}} \\
Short (1--2 calls)
& 234 & 1035 & 18.8 & 45.2 & +26.3
& [24.5, 28.4] & 60.9 & 0.5 \\
Medium (3--5 calls)
& 251 & 1149 & 14.2 & 47.3 & +33.1
& [30.9, 35.1] & 69.3 & 0.5 \\
Long ($\geq$6 calls)
& 165 & 1036 & 8.3 & 46.6 & +38.3
& [35.1, 41.3] & 82.0 & 0.7 \\

\midrule
\multicolumn{9}{l}{\textit{Number of application families}} \\
1
& 230 & 1167 & 18.8 & 51.2 & +32.4
& [30.3, 34.7] & 65.4 & 0.4 \\
2
& 360 & 1676 & 11.7 & 44.2 & +32.5
& [30.4, 34.8] & 72.1 & 0.6 \\
3+
& 55 & 352 & 5.9 & 39.4 & +33.4
& [28.5, 38.3] & 82.4 & 0.7 \\

\midrule
\multicolumn{9}{l}{\textit{Conditional branches}} \\
No conditional
& 303 & 1239 & 14.2 & 40.7 & +26.5
& [24.6, 28.3] & 66.5 & 0.6 \\
Has conditional
& 347 & 1981 & 13.6 & 50.0 & +36.4
& [34.3, 38.4] & 73.3 & 0.5 \\

\midrule
\multicolumn{9}{l}{\textit{Loops}} \\
No loop
& 511 & 2413 & 15.6 & 47.1 & +31.5
& [29.9, 33.1] & 67.9 & 0.4 \\
Has loop
& 139 & 807 & 8.5 & 44.3 & +35.7
& [32.5, 39.2] & 78.9 & 0.8 \\

\midrule
\multicolumn{9}{l}{\textit{Try/catch regions}} \\
No try/catch
& 413 & 1929 & 13.7 & 43.2 & +29.5
& [27.7, 31.4] & 67.9 & 0.6 \\
Has try/catch
& 237 & 1291 & 14.0 & 51.2 & +37.2
& [34.7, 39.6] & 74.8 & 0.5 \\

\midrule
\multicolumn{9}{l}{\textit{Control-flow composition}} \\
Linear
& 186 & 781 & 14.4 & 39.7 & +25.3
& [22.7, 28.0] & 64.2 & 0.5 \\
Branch only
& 132 & 688 & 14.9 & 48.6 & +33.6
& [30.2, 36.9] & 69.4 & 0.5 \\
Loop only
& 36 & 148 & 8.9 & 30.9 & +22.0
& [18.6, 26.2] & 70.4 & 0.7 \\
Error handling only
& 78 & 295 & 16.6 & 48.8 & +32.2
& [28.6, 36.1] & 69.6 & 0.5 \\
Multiple control-flow types
& 218 & 1308 & 12.8 & 50.4 & +37.7
& [35.2, 40.4] & 75.5 & 0.6 \\

\midrule
\multicolumn{9}{l}{\textit{Maximum nesting depth}} \\
1
& 186 & 781 & 14.4 & 39.7 & +25.3
& [22.9, 28.0] & 64.2 & 0.5 \\
2
& 235 & 1030 & 15.0 & 47.8 & +32.8
& [30.4, 35.3] & 69.3 & 0.4 \\
3+
& 229 & 1409 & 12.6 & 49.0 & +36.5
& [34.2, 38.8] & 75.3 & 0.6 \\

\midrule
\multicolumn{9}{l}{\textit{Fault injection}} \\
No injected fault
& 650 & 2956 & 13.6 & 45.8 & +32.2
& [30.7, 33.8] & 70.4 & 0.5 \\
Injected fault
& 224 & 264 & 16.4 & 52.8 & +36.5
& [33.8, 39.3] & 74.2 & 0.6 \\

\bottomrule
\end{tabular}
}
\end{table*}

\paragraph{The gap persists across workflow characteristics.}
As shown in Table~\ref{tab:workflow_characteristics},
\textsc{Step-by-Step} achieves higher Orchestration Accuracy than
\textsc{Global} in every reported subgroup. This pattern also holds at the
model level: all 45 evaluated models have a positive
\textsc{Step-by-Step}--\textsc{Global} gap in each reported subgroup.
Thus, the aggregate gap is not attributable to a single workflow-size,
application-count, or control-flow subgroup.

\paragraph{The gap increases with workflow size.}
The Orchestration Accuracy gap increases from 26.3 percentage points (pp)
for short workflows to 33.1 pp for medium workflows and 38.3 pp for long
workflows. This increase is driven primarily by lower \textsc{Global}
performance as workflow size increases: \textsc{Global} decreases from
18.8\% on short workflows to 8.3\% on long workflows, while
\textsc{Step-by-Step} remains between 45.2\% and 47.3\% across the three
size groups. The \textsc{Global} Error Rate also increases from 60.9\% to
82.0\% across these groups, whereas the \textsc{Step-by-Step} Error Rate
remains below 1\%.

\paragraph{Control-flow structure shows a similar pattern.}
Workflows containing conditional branches have a 36.4 pp gap, compared
with 26.5 pp for workflows without conditionals. The corresponding gaps are
35.7 pp versus 31.5 pp for workflows with and without loops, and 37.2 pp
versus 29.5 pp for workflows with and without try/catch regions.
When workflows are grouped by control-flow composition, linear workflows
have a 25.3 pp gap, while workflows containing multiple control-flow types
have a 37.7 pp gap. The same pattern appears with nesting depth: the gap is
25.3 pp at depth 1, 32.8 pp at depth 2, and 36.5 pp at depth 3 or greater.
Together, these stratifications show that the
\textsc{Step-by-Step}--\textsc{Global} gap is larger in workflows with
greater structural complexity, as measured by workflow size, control-flow
composition, and nesting depth.

\paragraph{Application count does not explain the gap.}
The gap is 32.4 pp for workflows using one application family, 32.5 pp for
two application families, and 33.4 pp for three or more application
families. Both protocols obtain lower Orchestration Accuracy as the number
of application families increases, but their difference remains similar
across these groups. Thus, the larger gaps observed for longer and more
structurally nested workflows are not reproduced by application count
alone.

\paragraph{The gap also persists under fault injection.}
For runtime variants without injected faults, the gap is 32.2 pp, compared
with 36.5 pp for fault-injected variants. Both settings therefore preserve
the main comparison between \textsc{Global} and \textsc{Step-by-Step}.
We do not interpret the higher absolute accuracy on fault-injected variants
as an effect of fault injection because the two variant groups differ in
the workflows and execution paths they contain.

Overall, these results show that the advantage of
\textsc{Step-by-Step} over \textsc{Global} is not concentrated in a
particular workflow subgroup. The gap is present across workflow sizes,
application counts, control-flow structures, nesting depths, and
fault-injection settings, and is larger under several measures of workflow
structural complexity.

\section{Inference Cost}
\label{app:inference_cost}

\textsc{Global} constructs a complete workflow before execution, whereas \textsc{Step-by-Step} repeatedly invokes the model as execution progresses. Table~\ref{tab:inference_cost} compares the resulting inference usage across open-weight and closed-source models.

\textbf{LLM calls.}
The reduction in repeated inference is consistent across model families. Across the models in Table~\ref{tab:inference_cost}, \textsc{Step-by-Step} uses $15.4\times$--$35.4\times$ more LLM calls than \textsc{Global}. For example, Qwen3-30B-A3B increases from 650 calls under \textsc{Global} to 22,997 under \textsc{Step-by-Step}, while Claude Opus 5 Thinking increases from 650 to 11,964 calls. On a per-variant basis, \textsc{Global} uses approximately 0.2 calls, compared with 3.1--7.1 calls for \textsc{Step-by-Step} among the models shown.

\textbf{Completion tokens.}
The larger number of calls also increases total generation. Across the models in Table~\ref{tab:inference_cost}, \textsc{Step-by-Step} generates $1.42\times$--$5.29\times$ more completion tokens than \textsc{Global}. For Qwen3.8-27B, completion-token usage increases from 1.07M under \textsc{Global} to 5.67M under \textsc{Step-by-Step}. Claude Opus 5 Thinking increases from 0.91M to 1.30M, while GPT-5.6 Sol Thinking increases from 0.59M to 1.40M.

Individual \textsc{Global} calls are substantially longer because each call generates a complete workflow. Qwen3.8-27B, for example, generates 1,652 completion tokens per call under \textsc{Global}, compared with 323 under \textsc{Step-by-Step}. This pattern holds broadly across the models in the table: \textsc{Global} uses fewer, longer calls, whereas \textsc{Step-by-Step} uses many more, shorter calls. The substantially larger number of \textsc{Step-by-Step} calls nevertheless results in higher total completion-token usage.

\textbf{Summary.}
\textsc{Global} substantially reduces inference usage: across the evaluated models, \textsc{Step-by-Step} requires $15.4\times$--$35.4\times$ more LLM calls and $1.42\times$--$5.29\times$ more completion tokens.
\begin{table*}[t]
\centering
\scriptsize
\renewcommand{\arraystretch}{1.05}
\setlength{\tabcolsep}{3.2pt}
\resizebox{\textwidth}{!}{
\begin{tabular}{l
                rr r
                rr
                rr r
                rr
                rr}
\toprule
&
\multicolumn{3}{c}{\textbf{Total LLM Calls}} &
\multicolumn{2}{c}{\textbf{Calls / Variant}} &
\multicolumn{3}{c}{\textbf{Completion Tokens (M)}} &
\multicolumn{2}{c}{\textbf{Tokens / Variant}} &
\multicolumn{2}{c}{\textbf{Tokens / Call}} \\
\cmidrule(lr){2-4}
\cmidrule(lr){5-6}
\cmidrule(lr){7-9}
\cmidrule(lr){10-11}
\cmidrule(lr){12-13}
\textbf{Model} &
\textsc{Global} &
\textsc{Step} &
\textbf{$\times$} &
\textsc{Global} &
\textsc{Step} &
\textsc{Global} &
\textsc{Step} &
\textbf{$\times$} &
\textsc{Global} &
\textsc{Step} &
\textsc{Global} &
\textsc{Step} \\
\midrule

\multicolumn{13}{l}{\textit{Open-weight models}} \\[1pt]

Gemma-3-12B
& 649 & 12,098 & \textbf{18.6$\times$}
& 0.20 & 3.76
& 0.54 & 1.92 & \textbf{3.55$\times$}
& 167.8 & 595.7
& 833 & 159 \\

Gemma-3-27B
& 649 & 14,253 & \textbf{22.0$\times$}
& 0.20 & 4.43
& 0.95 & 2.49 & \textbf{2.62$\times$}
& 295.8 & 774.3
& 1,468 & 175 \\

Gemma-4-12B
& 649 & 13,888 & \textbf{21.4$\times$}
& 0.20 & 4.31
& 1.18 & 3.29 & \textbf{2.79$\times$}
& 365.1 & 1,020.3
& 1,812 & 237 \\

Gemma-4-26B-A4B
& 649 & 17,198 & \textbf{26.5$\times$}
& 0.20 & 5.34
& 0.95 & 4.10 & \textbf{4.30$\times$}
& 296.2 & 1,272.9
& 1,470 & 238 \\

Gemma-4-31B
& 649 & 12,484 & \textbf{19.2$\times$}
& 0.20 & 3.88
& 1.16 & 2.89 & \textbf{2.48$\times$}
& 360.9 & 896.3
& 1,791 & 231 \\

GLM-4.5-Air
& 650 & 22,338 & \textbf{34.4$\times$}
& 0.20 & 6.94
& 0.85 & 4.21 & \textbf{4.98$\times$}
& 262.6 & 1,307.1
& 1,301 & 188 \\

GLM-4.5
& 650 & 15,488 & \textbf{23.8$\times$}
& 0.20 & 4.81
& 0.88 & 3.39 & \textbf{3.87$\times$}
& 272.0 & 1,053.7
& 1,348 & 219 \\

GLM-5.2
& 650 & 15,228 & \textbf{23.4$\times$}
& 0.20 & 4.73
& 0.81 & 2.55 & \textbf{3.13$\times$}
& 252.8 & 791.8
& 1,252 & 167 \\

Llama-3.3-70B
& 650 & 14,440 & \textbf{22.2$\times$}
& 0.20 & 4.48
& 0.72 & 1.59 & \textbf{2.23$\times$}
& 222.5 & 495.3
& 1,102 & 110 \\

Qwen2.5-72B
& 650 & 17,290 & \textbf{26.6$\times$}
& 0.20 & 5.37
& 0.99 & 2.21 & \textbf{2.24$\times$}
& 307.1 & 686.9
& 1,521 & 128 \\

Qwen3-30B-A3B
& 650 & 22,997 & \textbf{35.4$\times$}
& 0.20 & 7.14
& 1.13 & 3.71 & \textbf{3.28$\times$}
& 350.7 & 1,151.3
& 1,737 & 161 \\

Qwen3.8-27B
& 649 & 17,539 & \textbf{27.0$\times$}
& 0.20 & 5.45
& 1.07 & 5.67 & \textbf{5.29$\times$}
& 332.9 & 1,759.6
& 1,652 & 323 \\

\midrule
\multicolumn{13}{l}{\textit{Closed-source models}} \\[1pt]

GPT-3.5 Turbo
& 650 & 12,742 & \textbf{19.6$\times$}
& 0.20 & 3.96
& 0.45 & 1.93 & \textbf{4.30$\times$}
& 139.7 & 600.1
& 692 & 152 \\

GPT-4o
& 650 & 10,042 & \textbf{15.4$\times$}
& 0.20 & 3.12
& 0.68 & 1.08 & \textbf{1.60$\times$}
& 210.5 & 336.1
& 1,043 & 108 \\

Claude Opus 5 Thinking
& 650 & 11,964 & \textbf{18.4$\times$}
& 0.20 & 3.72
& 0.91 & 1.30 & \textbf{1.42$\times$}
& 284.0 & 402.9
& 1,407 & 108 \\

GPT-5.6 Sol Thinking
& 647 & 11,312 & \textbf{17.5$\times$}
& 0.20 & 3.51
& 0.59 & 1.40 & \textbf{2.37$\times$}
& 184.3 & 436.2
& 917 & 124 \\

\bottomrule
\end{tabular}
}
\caption{
\textbf{Inference usage under global and step-by-step orchestration.}
The $\times$ columns report \textsc{Step-by-Step}/\textsc{Global}.
For closed-source models, completion-token counts are estimated from retained
output text using median characters-per-token ratios measured from the
corresponding non-thinking runs; all other quantities are directly measured.
}
\label{tab:inference_cost}
\end{table*}

\section{Human Evaluation.}
\textbf{Source Recipes.}
\textsc{AgentOrch} is constructed from sanitized executable workflows instantiated in a Workato automation environment prior to benchmark construction. Customer-identifying information is removed before benchmark artifacts are derived from these workflows.

\textbf{Human Review of Task Descriptions.}
Four authors participated in an iterative manual validation process covering all benchmark task descriptions. In each round, task descriptions were reviewed against their corresponding source workflows for omitted workflow requirements, unsupported requirements, and disclosure of information excluded from the model’s task specification. Reviews were cross-checked across authors, and descriptions with identified issues were regenerated and subsequently re-evaluated. In the first review round, approximately 57\% of the task descriptions required regeneration. We repeated this review–regeneration–validation procedure for five rounds, until all task descriptions passed the review criteria.

\textbf{Execution-Based and Human Validation of Runtime Variants.}
For every generated runtime variant, we executed the corresponding reference workflow in the simulated environment and manually inspected the resulting execution behavior. 
The execution-based check verified that the runtime input and environment configuration supported the intended workflow execution, while human review examined the resulting tool outputs, state changes, control-flow behavior, and terminal outputs for consistency with the source workflow and intended runtime condition.
In the first validation round, approximately 38.5\% of generated runtime variants passed these checks.
Variants that failed either the execution-based or human validation criteria were regenerated and subsequently re-executed and re-reviewed.
This generate--execute--review cycle was repeated until all retained runtime variants passed the predefined validation criteria.
Because the benchmark infrastructure provides a local simulated execution environment with the same tool interfaces and benchmark-relevant connector semantics as the corresponding real connectors, we were able to inspect the execution outcome of every runtime variant rather than relying only on static input checks.

\section{Limitations}

\textbf{Scope.} \textsc{AgentOrch} targets structured enterprise workflows, where procedures are prescribed in advance and the same workflow is executed repeatedly on different inputs. This is the setting in which global orchestration is most practical. In GUI-based or open-ended environments, a complete executable workflow often cannot be specified before execution, and we leave the comparison in such settings to future work. Our workflows are derived from sanitized, executable Workato recipes spanning 64 tool interfaces across services such as Slack, Jira, Salesforce, and Snowflake; real enterprise workflows of this kind are rarely publicly available. Training and benchmark sets are disjoint by source recipe but drawn from the same enterprise automation environment. This mirrors deployment, where an organization trains on its existing workflows and applies the model to new workflows over the same tool ecosystem. The training set is also broader than the benchmark, covering 1,033 tool operations across 305 namespaces with only partial overlap with the benchmark tools. Transfer to unrelated automation domains remains to be studied.

\textbf{Protocol comparison.} We compare \textsc{Global} and \textsc{Step-by-Step} as deployed paradigms, so they differ in both when decisions are made and what information is available. Our additional protocols vary these factors separately: providing concrete runtime inputs alone leaves the gap largely unchanged, incremental construction adds step-wise decisions without execution information, and execution-feedback revision adds execution information without runtime decisions. A complete separation of the two factors is not possible, because using intermediate results at runtime without new decisions is precisely what a global workflow does through references between steps.

\textbf{Evaluation.} Our primary metric measures fidelity to the prescribed SOP, including its operation order, because enterprise procedures constrain how a task is carried out, not only its outcome. Final Accuracy, which evaluates only side effects and terminal output, complements this metric, and the gap and the gains from revision and training also hold under it. Model-generated workflows are executed in a simulated backend, since executing them against live enterprise services would produce irreversible side effects; reference-action replay shows high agreement between the simulator and the real connectors (Appendix~\ref{app:simulator_fidelity}).

\section{Full Results for Figure~\ref{fig:teaser}}
\label{app:full_results}
Table~\ref{tab:global_vs_react_all_models} reports the Orchestration Accuracy of all 31 models shown in Figure~\ref{fig:teaser} under \textsc{Global} orchestration and \textsc{Step-by-Step} interaction, from which the per-model gaps in the figure are computed.
\begin{table*}[t]
\centering
\caption{Orchestration accuracy under GLOBAL orchestration and
STEP-BY-STEP (ReAct) interaction. Results are reported as percentages.}
\label{tab:global_vs_react_all_models}
\resizebox{0.88\textwidth}{!}{
\begin{tabular}{lrr}
\toprule
\textbf{Model} &
\textbf{GLOBAL (\%)} &
\textbf{STEP-BY-STEP / ReAct (\%)} \\
\midrule
GPT-3.5 Turbo                         & 10.11 & 50.78 \\
GPT-4o                               & 28.09 & 50.12 \\

Llama 3.1 8B                         &  4.45 & 34.82 \\
Llama 3.1 70B                        & 10.81 & 38.92 \\
Mistral Nemo                         &  2.26 & 34.27 \\
Command R                            &  6.64 & 35.97 \\

Qwen 2.5 0.5B                        &  0.13 & 36.46 \\
Qwen 2.5 1.5B                        &  1.66 & 31.41 \\
Qwen 2.5 3B                          &  2.19 & 34.16 \\
Qwen 2.5 7B                          &  2.63 & 30.50 \\
Qwen 2.5 14B                         &  5.33 & 41.54 \\
Qwen 2.5 32B                         & 16.18 & 45.64 \\
Qwen 2.5 72B                         &  9.91 & 46.57 \\

Llama 3.2 1B                         &  0.00 & 36.93 \\
Llama 3.2 3B                         &  2.07 & 35.50 \\
Granite 3.1 2B                       &  6.30 & 31.83 \\
Granite 3.1 8B                       &  6.52 & 34.19 \\
Llama 3.3 70B                        & 10.45 & 42.98 \\

Gemma 3 4B                           &  6.15 & 32.75 \\
Gemma 3 12B                          &  8.46 & 37.85 \\
Gemma 3 27B                          &  7.41 & 39.93 \\
Mistral Small 3.1 24B                & 14.15 & 44.71 \\

Qwen 3 4B                            & 13.75 & 41.84 \\
Qwen 3 30B                           &  8.35 & 47.81 \\
GLM-4.5 Air FP8                      & 12.28 & 48.59 \\
GLM-4.5 FP8                          & 12.93 & 54.47 \\
DeepSeek V4 Flash                    & 27.01 & 58.99 \\
Qwen 3.8 27B FP8                     & 28.57 & 53.31 \\

Claude Opus 5                        & 46.44 & 63.52 \\
GPT-5.6 Sol Thinking                 & 51.76 & 66.88 \\
Claude Opus 5 Thinking               & 59.88 & 68.41 \\
\bottomrule
\end{tabular}
}
\end{table*}

\end{document}